%% file: main.tex
\documentclass{hld2026} 
\usepackage{graphicx}
\usepackage{subcaption}
\usepackage{amsmath,amssymb}
\usepackage{algorithm}
\usepackage{algorithmic}
\usepackage{booktabs}
\usepackage{hyperref}
\usepackage{url}

\title[Worker Disagreement Reveals Sharp Directions in Local SGD]{Worker Disagreement Reveals Sharp Directions in Local SGD}

\hldauthor{%
\Name{Tolga Dimlioglu} \Email{td2249@nyu.edu} \\
\Name{Kristi Topollai} \Email{kt2664@nyu.edu} \\
\Name{Anna Choromanska} \Email{ac5455@nyu.edu} \\
\addr {New York University, New York, USA} 
}

\begin{document}

\maketitle

\begin{abstract}
Deep neural network training often exhibits highly anisotropic loss geometry, where
a few sharp dominant Hessian directions coexist with a large flatter bulk.
Gradients tend to align disproportionately with these dominant directions, although stable progress
often requires movement through flatter bulk directions \cite{song2025does}.
Estimating the dominant subspace is therefore useful but costly with direct
Hessian-based methods. We show that standard Local SGD exposes this geometry
through worker disagreement. We theoretically show that the worker--average gap
covariance is shaped by stochastic-gradient noise and Hessian curvature, causing
workers to disagree along sharp, curvature-sensitive directions. Thus,
worker--average gaps provide a cheap Hessian-free estimator of the dominant
subspace. Experiments on MLPs, CNNs, and Transformers show that subspaces formed by worker--average gaps capture a substantial fraction of the gradient component lying in the dominant Hessian eigenspace.




\end{abstract}



\section{Introduction}


Deep neural networks are optimized in extremely high-dimensional parameter
spaces, yet their training dynamics often exhibit low-dimensional structure
\cite{9782552,yaras24compressible}. In particular, the training-loss Hessian is
highly anisotropic: a few large eigenvalues define a sharp low-dimensional
\emph{dominant} subspace, while the remaining directions form a flatter
high-dimensional \emph{bulk} \citep{sagun2016eigenvalues,sagun2017empirical,
papyan2018full,ghorbani2019investigation}. Gradients along centralized SGD
trajectories often concentrate in the dominant eigenspace
\citep{gur2018gradient,arous2024highdimsgd}, but recent evidence shows that useful
loss descent can depend substantially on movement through flatter bulk directions
\citep{song2025does}. Thus, identifying the dominant subspace is important for
diagnosing and controlling optimization, but direct Hessian-based estimation is
computationally expensive.

We investigate whether distributed training dynamics can reveal the
dominant subspace without explicitly estimating the Hessian. 
In Local SGD, workers independently take multiple stochastic-gradient steps before averaging \citep{stich2019local,mcmahan2017communication}, naturally \textit{disagreeing} on the next iterate. We quantify this disagreement via worker--average gaps.
Prior analyses typically treat these gaps as consensus error
to be bounded \citep{stich2019local,haddadpour2019local}. We instead ask whether
they reveal useful landscape directions.


We answer this question theoretically and empirically. We show that
worker--average gap covariance is shaped by the interaction between
stochastic-gradient noise and Hessian curvature, amplifying disagreement along
directions that are both noisy and curvature-sensitive. Since SGD noise is highly anisotropic and curvature-aligned
\citep{zhu2018anisotropic,Xie2023ontheoverlooked,
tang2025investigating,zhang2026superlinear}, worker--average gaps can serve as a
Hessian-free proxy for the dominant eigenspace. 
Experiments on MLPs, CNNs, and Transformers show that worker--average gap subspaces capture, and can be used to suppress, a substantial fraction of the gradient’s dominant Hessian component. These results recast worker disagreement from a consensus error into a low-cost source of geometric tool for controlling optimization.



\section{Preliminaries and Initial Observations}
\label{sec:setup}

\vspace{-1mm}
\paragraph{Setup.} We consider distributed empirical risk minimization with $M$ workers minimizing
$f(\theta)=\frac{1}{N}\sum_{n=1}^{N}\ell(h_\theta(x_n),y_n)$, where
$h_\theta:\mathcal{X}\to\mathcal{Y}$ is a neural network with parameters
$\theta\in\mathbb{R}^D$ and $\ell$ is the per-sample loss. The training set is
partitioned IID across $M$ workers. We focus on Local SGD training. At the beginning of each
communication round, all workers start from a common parameter vector; each worker
then performs $\tau$ local stochastic-gradient steps on its own minibatches; and
the workers synchronize by averaging their parameters. Pseudo-code is provided in Appendix~\ref{alg:local_sgd_gaps}.

\vspace{-1mm}
\paragraph{Notations.} Here, we follow the dominant--bulk subspace setup of~\citep{song2025does}. Let
$\nabla f(\theta)\in\mathbb{R}^D$ denote the full-batch gradient and
$H(\theta)=\nabla^2f(\theta)\in\mathbb{R}^{D\times D}$ denote the loss Hessian.
Let $\lambda_1(\theta)\geq\cdots\geq\lambda_D(\theta)$ be the eigenvalues of
$H(\theta)$, with corresponding eigenvectors $u_1(\theta),\ldots,u_D(\theta)$. 
For a chosen dimension $C$, we define the \emph{Dominant} subspace as
$\mathcal{S}_C(\theta)=\operatorname{span}\{u_1(\theta),\ldots,u_C(\theta)\}$,
and refer to its orthogonal complement $\mathcal{S}_C^\perp(\theta)$ as the \emph{Bulk}
subspace. Following~\citep{gur2018gradient, song2025does}, we set $C$ to the number of classes in the classification task. We denote the orthogonal projector onto
$\mathcal{S}_C(\theta)$ by $P_C(\theta)=\sum_{k=1}^C u_k(\theta)u_k(\theta)^\top$, and define
$P_C^\perp(\theta)=I-P_C(\theta)$.

For any vector $v\in\mathbb{R}^D$, we measure its relative alignment with the
dominant subspace using
\vspace{-1.1mm}
\begin{equation}
    \chi_C(v; \theta) := \| P_C (\theta) v \|_2 / \| v \|_2
    \label{eq:dominant_alignment}
    \vspace{-1.4mm} 
\end{equation}

This metric is inherited from \citet{song2025does}. Notice that, when $\chi_C(v;\theta)$ is close to one, we say $v$ is aligned with $\mathcal{S}_C(\theta)$ as most of the norm of $v$ lies in the dominant subspace; when it is close to zero, $v$ lies mostly in the bulk subspace. Next, we verify that the dominant-subspace alignment observed in centralized SGD also appears along Local SGD trajectories.

\begin{figure}[t]
    \vspace{-2mm}
    \centering
    \begin{minipage}[b]{0.32\textwidth}
        \centering
        \includegraphics[width=\textwidth]{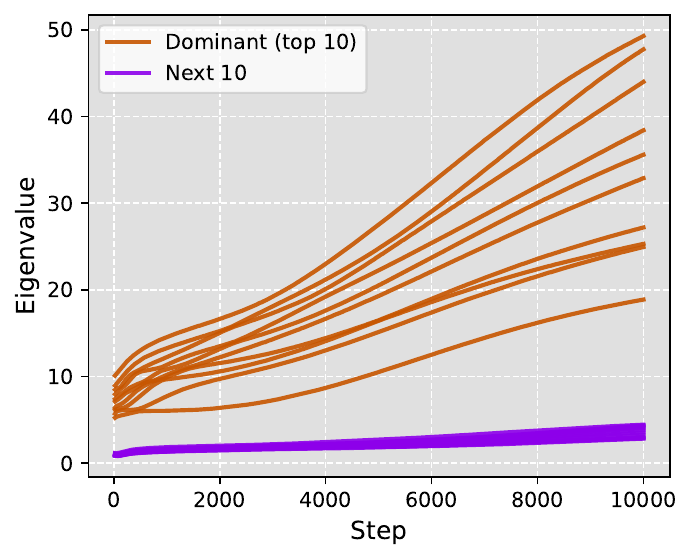}
    \end{minipage}
    \hfill
    \begin{minipage}[b]{0.32\textwidth}
        \centering
        \includegraphics[width=\textwidth]{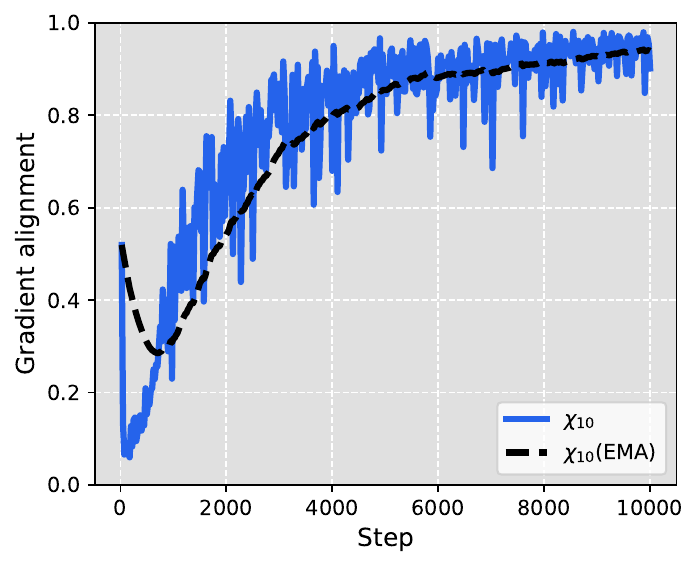}
    \end{minipage}
    \hfill
    \begin{minipage}[b]{0.34\textwidth}
        \centering
        \includegraphics[width=\textwidth]{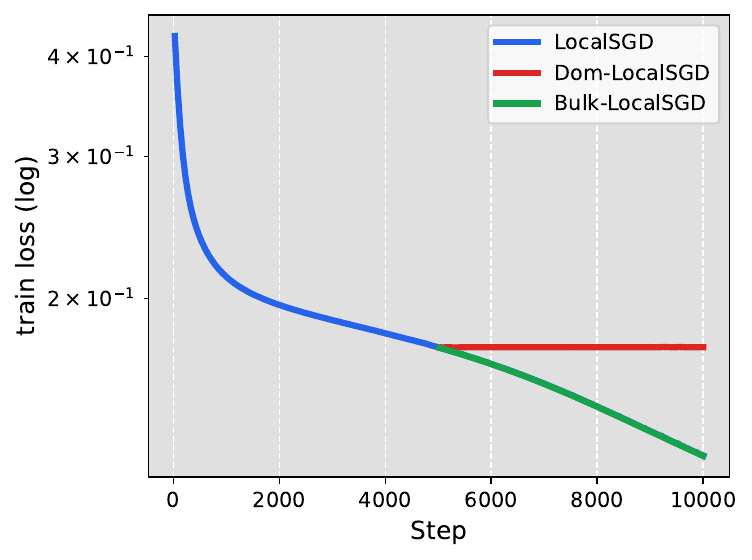}  
    \end{minipage}
\vspace{-2mm}   
\caption{
FC tanh network trained on MNIST-5k using Local SGD with $M=4$ workers and $\tau=5$. 
\textbf{Left:} the Hessian spectrum develops a clear separation between the top $10$ eigenvalues and the bulk. 
\textbf{Middle:} the gradient direction increasingly aligns with the dominant Hessian subspace, measured by $\chi_{10}$.
\textbf{Right:} retaining only the dominant component stalls optimization, indicating that useful descent is largely carried by bulk directions.
}
\label{fig:mnist_initial_observations}
\vspace{-3mm}
\end{figure}



\vspace{-1mm}
\paragraph{Observation 1: Local SGD gradients align with the dominant subspace.}
At each synchronization round, after worker averaging, we compute the full-batch gradient $\nabla f(\bar \theta)$ and
Hessian $H(\bar \theta)$ at the averaged parameter $\bar \theta$, form the dominant
subspace $\mathcal{S}_C(\bar \theta)$, and report
$\chi_C(\nabla f(\bar \theta);\bar \theta)$. Figure~\ref{fig:mnist_initial_observations}
shows the result for a fully-connected (FC) tanh network trained on MNIST \cite{mnist} with $M=4$ workers and
communication period $\tau=5$. The Hessian develops a clearly separated dominant
subspace, and the gradient increasingly aligns with this subspace throughout
training as shown in left and middle panels of  Figure~\ref{fig:mnist_initial_observations} respectively. We observe the same qualitative behavior for a CNN on CIFAR10 \cite{cifar10} and a
Transformer on SST2 \cite{sst2}; but these results are deferred to Appendix~\ref{app:subsec:dom_bulk_stage} due to space constraints. 
Then, we test whether this dominant-space alignment implies that Local SGD can be 
continued using only the dominant component of its communication-round update.


\vspace{-2mm}
\paragraph{Observation 2: Local SGD progress is not carried by the dominant component.}
Starting from a standard Local SGD checkpoint, we compare standard Local SGD with
two projected variants. Let
$\bar p^c:= \frac{1}{M}\sum_{i=1}^M \theta_i^{c,\tau}-\bar \theta^c$
denote the average outer step at communication round $c$. Dom-Local SGD projects
$\bar p^c$ onto the dominant subspace $\mathcal{S}_C(\bar \theta^c)$, whereas
Bulk-Local SGD projects it onto the complementary bulk subspace
$\mathcal{S}_C^\perp(\bar \theta^c)$. As shown in
Figure~\ref{fig:mnist_initial_observations} right panel, Dom-Local SGD stalls after
the projection is introduced, whereas Bulk-Local SGD closely tracks standard Local
SGD. Thus, although gradients are strongly aligned with the dominant Hessian
subspace, the update component that sustains training lies primarily in the bulk.
More details and results on CIFAR-10 and SST-2 are given in
Appendix~\ref{app:subsec:dom_bulk_stage}.
Next, we look beyond the average outer step of Local SGD and analyze the
disagreement among local worker parameters, i.e., the gap vector between each worker
parameter and their average, which Local SGD produces naturally during training.




\vspace{-2mm}
\section{Theoretical Characterization of Worker--Average Gaps}
\label{sec:gap_theory}
\vspace{-1mm}
We analyze how worker disagreements, i.e., the worker--average gaps, arise within communication round \(c\). At the beginning of the round, all
workers are synchronized at \(\bar\theta^c\), so that
\( \theta_i^{c,0}=\bar\theta^c \) for \( i=1,\ldots,M\).
Each worker then performs \(\tau\) local SGD steps with step size \(\eta\):
\vspace{-0.5mm}
\begin{equation}
    \theta_i^{c,t}
    =
    \theta_i^{c,t-1}
    -
    \eta g_i^{c,t},
    \qquad
    g_i^{c,t}
    =
    \nabla f(\theta_i^{c,t-1})+\epsilon_i^{c,t},
    \qquad
    t=1,\ldots,\tau ,
    \label{eq:local_sgd_update_theory}
\end{equation}

\vspace{-1mm} where \(\epsilon_i^{c,t}\) is the stochastic-gradient noise satisfying $\mathbb E[\epsilon_i^{c,t}\mid \theta_i^{c,t-1}]=0$. Now, define the within-round average and the worker--average gap \(i\) from this
average as $ \bar\theta^{c,t} := \frac{1}{M}\sum_{j=1}^{M}\theta_j^{c,t}$ and $ d_i^{c,t} := \theta_i^{c,t}-\bar\theta^{c,t}$ respectively. Notice that, by construction, \(\sum_i d_i^{c,t}=0\), and since workers are synchronized at
the start of the round, \(d_i^{c,0}=0\). After the final local step, the next 
synchronized model is $\bar\theta^{c+1}:=\bar\theta^{c,\tau}$ and the worker--average gap is
\(
    \Delta_i^{c+1}
    :=
    \theta_i^{c,\tau}-\bar\theta^{c+1}
    =
    d_i^{c,\tau}.
    \label{eq:worker_average_gap_definition}
\)
The following lemma describes the within-round evolution of the worker-average gaps.

\begin{lemma}[Linearized worker--average gap dynamics]
\label{lem:gap_recurrence}
Assume that \(f\) is twice differentiable. For each local step \(t\ \in \{1, 2, \dots, \tau \} \) in
communication round \(c\), define the local Hessian around the within-round
average by $H_{c,t}:=\nabla^2 f(\bar\theta^{c,t-1})$.
Let $ \bar\epsilon^{c,t} :=  \frac{1}{M}\sum_{j=1}^{M}\epsilon_j^{c,t}$ and $ \zeta_i^{c,t} := \epsilon_i^{c,t}-\bar\epsilon^{c,t}.$ Then the within-round worker--average gap approximately satisfies
\vspace{-1mm}
\begin{equation}
    d_i^{c,t}
    \approx
    (I-\eta H_{c,t})d_i^{c,t-1}
    -
    \eta\zeta_i^{c,t}.
    \label{eq:gap_recurrence}
\end{equation}
\end{lemma}
\vspace{-1mm} Lemma~\ref{lem:gap_recurrence} shows that worker-average gaps are driven by 
centered stochastic-gradient noise, while the local Hessian determines how 
previous gaps propagate through the within-round dynamics.

\begin{theorem}[Worker--average gap covariance as propagated stochastic noise]
\label{thm:gap_covariance}
Under the approximation in Lemma~\ref{lem:gap_recurrence}, suppose that the
Hessian does not change substantially within one communication round, so that $H_{c,t}\approx H_c$ for $t=1,\ldots,\tau$. Then the final worker--average gap at the end of communication round \(c\)
satisfies
\vspace{-3mm}
\begin{equation}
    \Delta_i^{c+1}
    =
    d_i^{c,\tau}
    \approx
    -\eta
    \sum_{t=1}^{\tau}
    (I-\eta H_c)^{\tau-t}
    \zeta_i^{c,t}
    \label{eq:gap_unrolled_constant_hessian}
\end{equation}

\vspace{-1.5mm}Assume further that the stochastic-gradient noise is independent across workers
and local steps, with $\operatorname{Cov}(\epsilon_i^{c,t})=\Sigma_c$.
Then the centered worker noise satisfies
\( \operatorname{Cov}(\zeta_i^{c,t}) =\left(1-\frac{1}{M}\right)\Sigma_c\)
and the worker-gap covariance is
\vspace{-1mm}
\begin{equation}
    \boxed{
    \operatorname{Cov}(\Delta_i^{c+1})
    \approx
    \eta^2
    \left(1-\frac{1}{M}\right)
    \sum_{q=0}^{\tau-1}
    (I-\eta H_c)^q
    \Sigma_c
    (I-\eta H_c)^q
    }
    \label{eq:gap_covariance_curvature_derivation}
\end{equation}
\end{theorem}

\vspace{-1mm}The takeaway from Theorem~\ref{thm:gap_covariance} is that the covariance
of worker-average gaps at the end of the communication round, is shaped jointly by the stochastic-gradient noise covariance
\(\Sigma_c\) and the local curvature \(H_c\). We next project
this covariance onto individual Hessian eigendirections to characterize where
the worker disagreement is largest in the eigenbasis.

\vspace{-2mm}
\begin{proposition}[Directional gap variance under noise--curvature coupling]
\label{prop:noise_curvature}
Let \( H_c=U_c\Lambda_c U_c^\top \) and \( \Lambda_c=\operatorname{diag}(\lambda_{1,c},\ldots,\lambda_{D,c}), \)
where \(u_{r,c}\), the \(r\)-th column of \(U_c\), is the Hessian eigenvector
corresponding to eigenvalue \(\lambda_{r,c}\). Suppose that the
stochastic-gradient noise covariance is approximately diagonal in the Hessian
eigenbasis: \(
    U_c^\top \Sigma_c U_c
    \approx
    \operatorname{diag}
    \left(
    \sigma_{1,c}^2,\ldots,\sigma_{D,c}^2
    \right),\)
where \(\sigma_{r,c}^2\) denotes the noise variance along Hessian eigendirection
\(u_{r,c}\). Then
\vspace{-3mm}
\begin{equation}
    \operatorname{Var}
    \left(
    \langle \Delta_i^{c+1},u_{r,c}\rangle
    \right)
    \approx
    \eta^2
    \left(1-\frac{1}{M}\right)
    \sigma_{r,c}^2
    \sum_{q=0}^{\tau-1}(1-\eta\lambda_{r,c})^{2q}.
    \label{eq:gap_variance_eigendirection}
\end{equation}

\vspace{-3mm}We write \(
    \operatorname{Var}
    \left(
    \langle \Delta_i^{c+1},u_{r,c}\rangle
    \right)
    \approx
    \eta^2
    \left(1-\frac{1}{M}\right)
    \sigma_{r,c}^2
    \psi_\tau(\eta\lambda_{r,c})
    \label{eq:gap_variance_filter}
\)
where $\psi_\tau(a):=\sum_{q=0}^{\tau-1}(1-a)^{2q}$.
Furthermore, if the noise variance obeys the empirically supported
noise--curvature scaling $\sigma_{r,c}^2 \propto  \lambda_{r,c}^{\gamma}$ 
where $ 1\leq \gamma\leq 2$ \cite{zhang2026superlinear}, then
\vspace{-2.2mm}
\begin{equation}
    \operatorname{Var}
    \left(
    \langle \Delta_i^{c+1},u_{r,c}\rangle
    \right)
    \propto
    \lambda_{r,c}^{\gamma}
    \psi_\tau(\eta\lambda_{r,c}).
    \label{eq:gap_variance_noise_curvature_scaling}
\end{equation}
\end{proposition}
\vspace{-2.5mm}
Proposition~\ref{prop:noise_curvature} links worker disagreement
to Hessian curvature at the eigendirection level. The term
\(\sigma_{r,c}^2\) captures how much stochastic-gradient noise is injected
along \(u_{r,c}\), while \(\psi_\tau(\eta\lambda_{r,c})\) captures how local
SGD propagates that noise before synchronization. Recent evidence suggests
that \(\sigma_{r,c}^2\) itself grows with curvature, approximately as
\(\lambda_{r,c}^{\gamma}\) where $ 1\leq \gamma\leq 2$ \cite{zhang2026superlinear}. 
Under this coupling, worker--average gaps tend to become large along
high-curvature Hessian directions. Since the dominant subspace is precisely the
span of the leading Hessian eigendirections, i.e., the directions with the
largest eigenvalues, this suggests that the span of worker--average gaps can be
used as a data-driven estimator of the dominant subspace. 

The detailed proofs of the theory presented here can be found in Section~\ref{app:full_theory} of the
Appendix.

\vspace{-2mm}
\section{Worker--Average Gaps as a Dominant Subspace Estimator}
\label{sec:main_empirical_results}
\vspace{-1mm}
We test our theoretical claims in Section~\ref{sec:gap_theory}  by constructing a subspace from observed worker gaps during standard Local SGD and measuring its alignment with the top Hessian eigenspace.
To form the gap-based proxy subspace, we maintain a FIFO buffer of
worker--average gaps collected during standard Local SGD training. At
synchronization round \(c\), we insert the observed gaps
\(\{\Delta_i^c\}_{i=1}^{M-1}\) into the buffer (Not $M$ due to linear dependence).
We denote the current buffer by \( Z_c=[z_1,\ldots,z_B]\in\mathbb{R}^{D\times B}\)
where each buffer entry \(z_j\) is a previously observed worker--average gap
\(\Delta_i^{c}\), \(B\) is the buffer capacity, and \(D\) is the number of model
parameters. We construct the gap subspace from the Gram matrix \(G_c=Z_c^\top Z_c\). Let
\(G_c=V_c \Omega_c V_c^\top\) be its eigendecomposition. We form the
orthonormal basis \(Q_c = Z_c V_c \Omega_c ^{-1/2}\). Then
\(Q_c\in\mathbb{R}^{D\times B_c}\) ($B_c$ is the retained rank) has orthonormal columns and spans the column
space of the gap buffer. Additional implementation details are given in
Appendix~\ref{app:subsec:gap_subspace}.




Recall that \(P_C\) denotes the orthogonal projector onto the dominant Hessian
subspace, obtained from the top-\(C\) Hessian eigendirections. Similarly, we define
\(P_{Q_c}=Q_cQ_c^\top\) as the projector obtained from the
worker--average gap subspace. For a vector \(v\), we measure how much of its true
dominant component is removed by the proxy filter \(I-P_{Q_c}\) using
\begin{equation}
    \rho_c(v) := 1 -  \frac{\left\|P_C(I-P_{Q_c})v\right\|_2} {\left\|P_Cv\right\|_2}.
    \label{eq:dom_mass_removed_frac}
\end{equation}
 Thus, \(\rho_c(v)\) measures the fraction of the true
dominant component of \(v\) suppressed by the worker-gap proxy subspace $Q_c$. Larger
values indicate that the gap subspace more effectively captures the
Hessian-based dominant directions.
In our experiments, we evaluate this quantity on the full-batch gradient
\(v=\nabla f(\bar \theta^c)\) at the averaged model \(\bar \theta^c\) at
communication round \(c\), similar to the setup described in \textit{Observation 2} of Section~\ref{sec:setup}.

\begin{figure}[t]
    \centering
    \begin{minipage}[b]{0.325\textwidth}
        \centering
        \includegraphics[width=\textwidth]{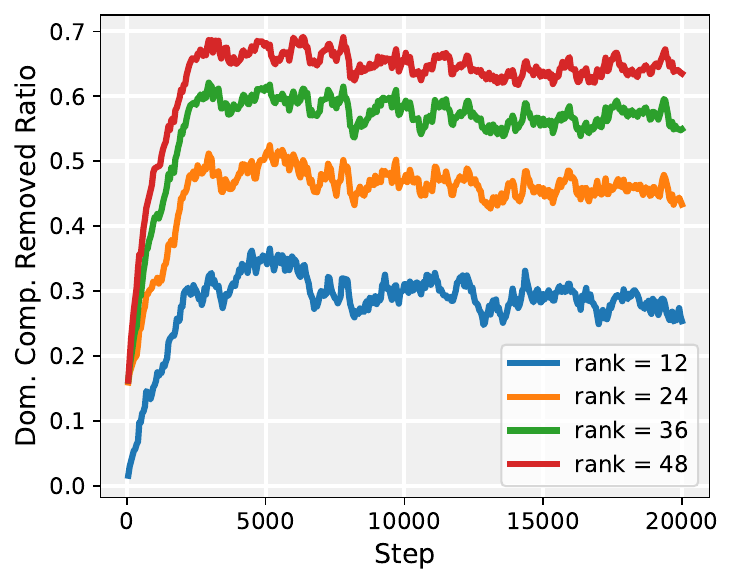}
        \setlength{\abovecaptionskip}{-5mm}
        \caption*{(a) FC on MNIST-5k}
    \end{minipage}
    \hfill
    \begin{minipage}[b]{0.325\textwidth}
        \centering
        \includegraphics[width=\textwidth]{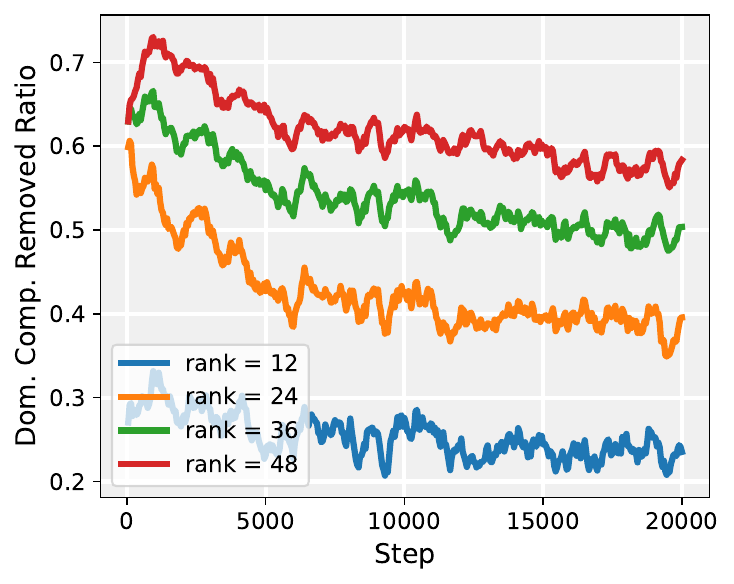}
        \setlength{\abovecaptionskip}{-5mm}
        \caption*{(b) CNN on CIFAR10-5k}
    \end{minipage}
    \hfill
    \begin{minipage}[b]{0.325\textwidth}
        \centering
        \includegraphics[width=\textwidth]{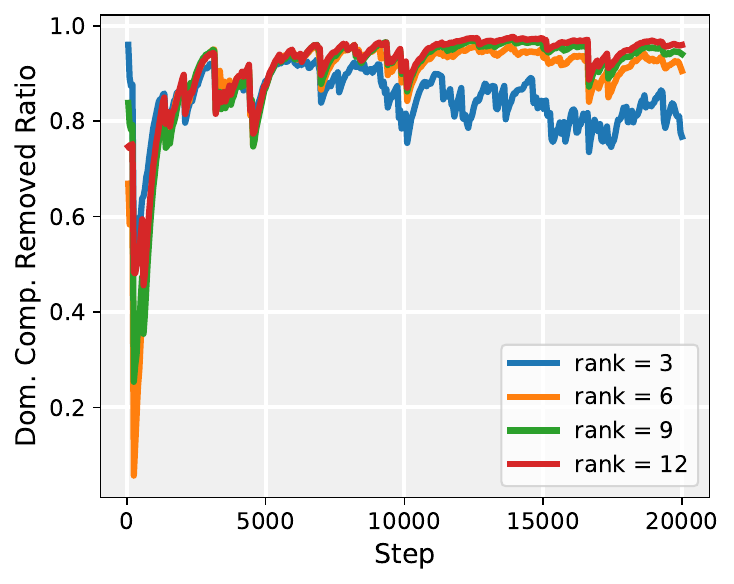} 
        \setlength{\abovecaptionskip}{-5mm}
        \caption*{(c) Transformer on SST2-5k}
    \end{minipage}
    \vspace{-2mm}
    \caption{Dominant-component removal fraction achieved by worker-gap subspaces of
different buffer capacities. The worker-gap subspace captures a large portion of
the true dominant Hessian component across architectures, with coverage improving
as the buffer capacity increases. (Curves are shown with exponential moving
average smoothing for clarity.)}
\vspace{-6mm}
\label{fig:gap_subspace_results}  
\end{figure}

To assess effectiveness across architectures and data modalities, we run experiments
on three model--dataset pairs: a tanh FC network on MNIST, a ReLU
CNN on CIFAR10, and a 2-layer Transformer on SST2. For each dataset, we train
on a subset of 5k samples using standard Local SGD with \(M=4\) workers and
communication period \(\tau=5\). Additional experimental details are provided in
Appendix~\ref{app:subsec:gap_subspace}. Following \cite{gur2018gradient,song2025does}, we set \(C=10\) for MNIST and CIFAR10, and \(C=2\) for SST2. Accordingly, for MNIST and CIFAR10, we sweep
FIFO buffer capacities \(B\in\{12,24,36,48\}\). For SST2, we sweep
\(B\in\{3,6,9,12\}\). Results are shared in Figure~\ref{fig:gap_subspace_results}.

Figure~\ref{fig:gap_subspace_results} shows that the worker-gap subspace removes
a substantial fraction of the gradient's dominant Hessian component across all
three settings. Larger buffer capacities consistently improve this removal,
showing that having more recent worker--average gaps capture more of the dominant
eigenspace. The effect is strongest for SST2, where even small buffers remove
most of the dominant component, while MNIST and CIFAR10 improve more gradually
with rank. Overall, these results support our theory that worker disagreement
concentrates along sharp, curvature-sensitive directions, making
worker--average gaps an effective Hessian-free proxy for the dominant subspace.

\vspace{-2mm}
\section{Conclusion and Future Work}
\label{sec:conclusion}
\vspace{-1mm}
In this work, we showed that the worker--average gaps naturally produced by
Local SGD carry useful information about sharp directions in the loss landscape.
Our theory links worker--average gaps to stochastic-gradient noise and Hessian
curvature, showing that under a noise--curvature coupling
\cite{zhang2026superlinear}, gaps become large along high-curvature Hessian
directions. This motivates using their span as a Hessian-free dominant-subspace
estimator, which empirically captures a substantial fraction of the dominant
Hessian component across model--dataset pairs. Appendix~\ref{app:sec:additional_results}
provides additional results on the effect of $\tau$ on the quality of the
gap-based subspace; together with preliminary filtering experiments showing that
suppressing the gap-estimated dominant component and amplifying its orthogonal
complement can accelerate optimization. We view the design of optimization
mechanisms that leverage worker--average gaps as a promising direction for
future work.

\bibliography{sample}
\newpage
\clearpage
\appendix
\input{appendix}

\end{document}

%% file: appendix.tex

\section{Experiment Details and More Results}
\label{app:sec:experiment_details}

Here, we provide more details of the experimental setup. To run the experiments, we use Python 3.11.15 programming language, PyTorch 2.8.0 framework with torchvision 0.23.0 and CUDA 12.6.
We used 3 machines, each equipped with 4 $\times$ GTX 1080 GPUs to run the experiments.

\subsection{Local SGD Training}

We consider distributed empirical risk minimization with $M$ workers minimizing
\[
    f(\theta)
    =
    \frac{1}{N}
    \sum_{n=1}^{N}
    \ell(h_\theta(x_n),y_n),
\]
where $h_\theta:\mathcal{X}\to\mathcal{Y}$ is a neural network with parameters
$\theta\in\mathbb{R}^D$, and $\ell$ denotes the per-sample loss. The training set
is partitioned IID across the $M$ workers. We denote the local dataset on worker
$i$ by $\mathcal{D}_i$, so that, for equal-size partitions,
\[
    f(\theta)
    =
    \frac{1}{M}
    \sum_{i=1}^{M}
    f_i(\theta),
    \qquad
    f_i(\theta)
    =
    \frac{1}{|\mathcal{D}_i|}
    \sum_{(x,y)\in\mathcal{D}_i}
    \ell(h_\theta(x),y).
\]
Local SGD proceeds in communication rounds. We use $c$ to index communication
rounds and $s$ to index local steps within a communication round. At the
beginning of communication round $c$, all workers are initialized from the same
averaged model $\bar{\theta}^{c}$. Each worker then independently performs
$\tau$ stochastic-gradient steps on minibatches sampled from its local data:
\[
    \theta_i^{c,s+1}
    =
    \theta_i^{c,s}
    -
    \eta
    \nabla f_{i,\mathcal{B}_i^{c,s}}
    \left(\theta_i^{c,s}\right),
    \qquad
    s=0,\ldots,\tau-1,
\]
where $\mathcal{B}_i^{c,s}$ is the minibatch sampled by worker $i$ at local step
$s$ of communication round $c$. After $\tau$ local steps, the workers synchronize
by averaging their local parameters:
\[
    \bar{\theta}^{c+1}
    =
    \frac{1}{M}
    \sum_{i=1}^{M}
    \theta_i^{c,\tau}.
\]
Thus, compared with fully synchronized minibatch SGD, Local SGD communicates only
once every $\tau$ local updates which reduces the time spent on communication by a factor of $\tau$. 
In our experiments, we also record the worker--average gap vectors before synchronization,
\[
    \Delta_i^{c+1}
    =
    \theta_i^{c,\tau}
    -
    \bar{\theta}^{c+1},
    \qquad i=1,\ldots,M,
\]
which quantify the disagreement accumulated by the local workers during the
communication period. These gaps are the basic objects used to construct the
gap-based subspace estimator analyzed in the main text. Pseudo-code is provided
in Algorithm~\ref{alg:local_sgd_gaps}.

\begin{algorithm}[H]
\caption{Local SGD with Worker--Average Gaps}
\label{alg:local_sgd_gaps}
\begin{algorithmic}[1]
\REQUIRE Number of workers $M$, communication period $\tau$, number of communication rounds $C$, learning rate $\eta$
\STATE Initialize global model $\bar{\theta}^{0}$
\FOR{$c=0,1,\ldots,C-1$}
    \FOR{each worker $i=1,\ldots,M$ in parallel}
        \STATE Set $\theta_{i}^{c,0} \leftarrow \bar{\theta}^{c}$
        \FOR{$s=0,1,\ldots,\tau-1$}
            \STATE Sample minibatch $\mathcal{B}_{i}^{c,s}$
            \STATE $g_{i}^{c,s} \leftarrow
            \nabla_\theta f_{\mathcal{B}_{i}^{c,s}}
            \left(\theta_{i}^{c,s}\right)$
            \STATE $\theta_{i}^{c,s+1}
            \leftarrow
            \theta_{i}^{c,s}-\eta g_{i}^{c,s}$
        \ENDFOR
    \ENDFOR
    \STATE Compute the worker average
    \(
        \bar{\theta}^{c+1}
        =
        \frac{1}{M}
        \sum_{i=1}^{M}
        \theta_{i}^{c,\tau}
    \)
    \STATE Compute worker--average gaps
    \(
        \Delta_i^{c+1}
        =
        \theta_i^{c,\tau}
        -
        \bar{\theta}^{c+1},
        \qquad i=1,\ldots,M.
    \)
\ENDFOR
\RETURN $\bar{\theta}^{C}$
\end{algorithmic}
\end{algorithm}

\subsection{Dominant--Bulk Subspace Phenomenon in Local SGD Training}
\label{app:subsec:dom_bulk_stage}

\paragraph{Hyperparameters.}
We train all models using Local SGD with $M=4$ workers and communication period
$\tau=5$, i.e., each worker performs five local SGD steps between consecutive
synchronizations. Training is run for 10,000 local steps, corresponding to 2,000
communication rounds. The local optimizer is vanilla SGD without momentum, and
we do not use weight decay. The per-worker batch size is set to 50. For each
dataset, we randomly select 5,000 training samples, which we denote by
MNIST-5k, CIFAR10-5k, and SST2-5k. We use mean squared error (MSE) as the
training loss $f$ for all experiments. We follow the experimental protocol of \cite{song2025does}, which covers
multiple architectures and data modalities. Below, we provide the architectural
details of the models used in our experiments.

\textbf{FC tanh:}
We train a fully connected neural network with two hidden layers and tanh
activations on MNIST-5k. Each hidden layer has width 200. Thus, the three linear
weight matrices have sizes
$[\texttt{input\_size},200]$, $[200,200]$, and
$[200,\texttt{output\_size}]$, with corresponding bias vectors of sizes
$[200]$, $[200]$, and $[\texttt{output\_size}]$. Since the model is trained on
MNIST, $\texttt{output\_size}=10$.

\textbf{CNN ReLU:}
We train a convolutional neural network with ReLU activations on CIFAR10-5k. The
network consists of two convolutional blocks, each using 32 output channels. Each
block applies a $3\times 3$ convolution with stride $1$ and padding $1$, followed
by a ReLU activation and $2\times 2$ max pooling. After the two convolutional
blocks, the feature map is flattened and passed to a linear classifier. Since the
model is trained on CIFAR10, the output dimension is set to 10.

\textbf{Transformer:}
For SST2-5k, we train a small Transformer encoder for binary classification. The
model uses token and positional embeddings with hidden dimension 64. The maximum
sequence length is set to the dataset-specific value used in preprocessing. The
encoder has two Transformer layers, each with two attention heads. Each layer
contains a multi-head self-attention block followed by a residual connection and
LayerNorm, and a feed-forward block consisting of two linear layers with a ReLU
activation between them, again followed by a residual connection and LayerNorm.
No dropout is used. The final sequence representation is obtained by mean pooling
over the sequence dimension, and a linear classifier maps the pooled
representation to two output classes. The classifier head is initialized with
zero weights and zero bias.

The learning rate is set to  $\eta=0.05$ for FC tanh and it is set to  $\eta=0.005$
for CNN ReLU and Transformer experiments.


To obtain the leading eigenvalue--eigenvector pairs of the Hessian, we use the
Lanczos algorithm. The Hessian is evaluated on the full training subset, i.e.,
using all 5,000 samples from the corresponding dataset. We compute the Hessian
spectrum and collect the relevant training statistics every 25 local steps,
which corresponds to every 5 communication rounds since $\tau=5$. Unless stated
otherwise, both the leading Hessian eigenvalue--eigenvector pairs and the
full-batch gradients are evaluated at the synchronized model
$\bar{\theta}^{c}$ at the beginning of a communication round, before the workers
perform their local updates. In other words, these quantities are computed at
the common parameter vector obtained from the most recent synchronization, prior
to the subsequent local exploration phase.

\begin{figure}[H]
    \centering
    \begin{minipage}[b]{0.325\textwidth}
        \centering
        \includegraphics[width=\textwidth]{figures/stage/FC-TANH/eigenvalues.pdf}
        \caption*{(a) FC on MNIST-5k}
    \end{minipage}
    \hfill
    \begin{minipage}[b]{0.325\textwidth}
        \centering
        \includegraphics[width=\textwidth]{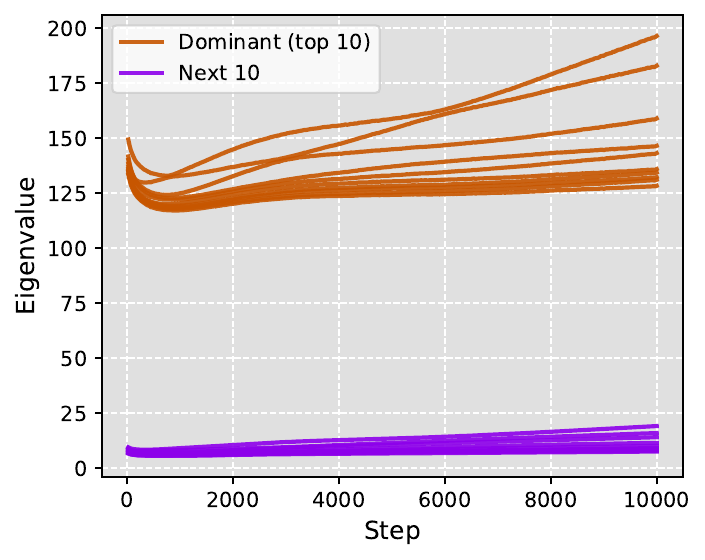}
        \caption*{(b) CNN on CIFAR10-5k}
    \end{minipage}
    \hfill
    \begin{minipage}[b]{0.325\textwidth}
        \centering
        \includegraphics[width=\textwidth]{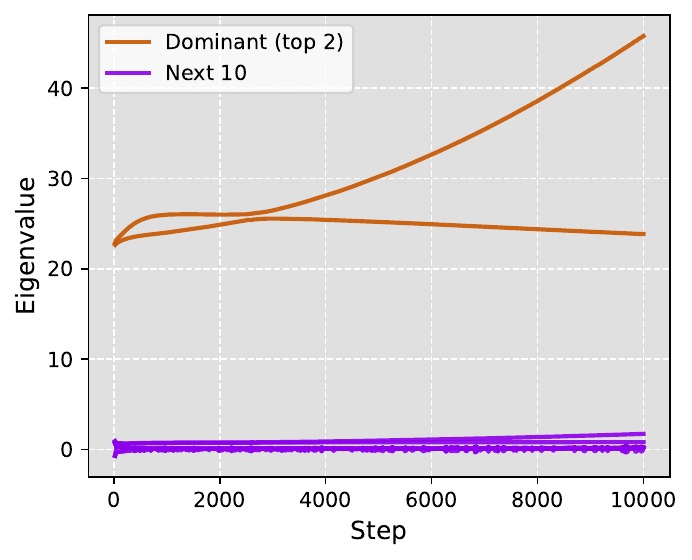} 
        \caption*{(c) Transformers on SST2-5k}
    \end{minipage}
    \caption{Evolution of Hessian eigenvalues during Local SGD training. Across all
three settings, a small number of dominant eigenvalues separates from the
remaining spectrum, confirming the dominant--bulk Hessian structure.}
    \label{app:fig:eigenvalues}
\end{figure}

\begin{figure}[H]
    \centering
    \begin{minipage}[b]{0.325\textwidth}
        \centering
        \includegraphics[width=\textwidth]{figures/stage/FC-TANH/grad_alignment.pdf}
        \caption*{(a) FC on MNIST-5k}
    \end{minipage}
    \hfill
    \begin{minipage}[b]{0.325\textwidth}
        \centering
        \includegraphics[width=\textwidth]{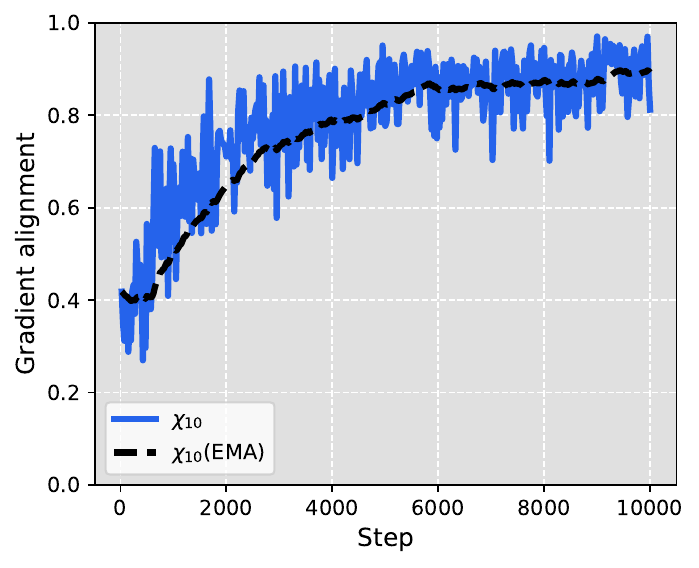}
        \caption*{(b) CNN on CIFAR10-5k}
    \end{minipage}
    \hfill
    \begin{minipage}[b]{0.325\textwidth}
        \centering
        \includegraphics[width=\textwidth]{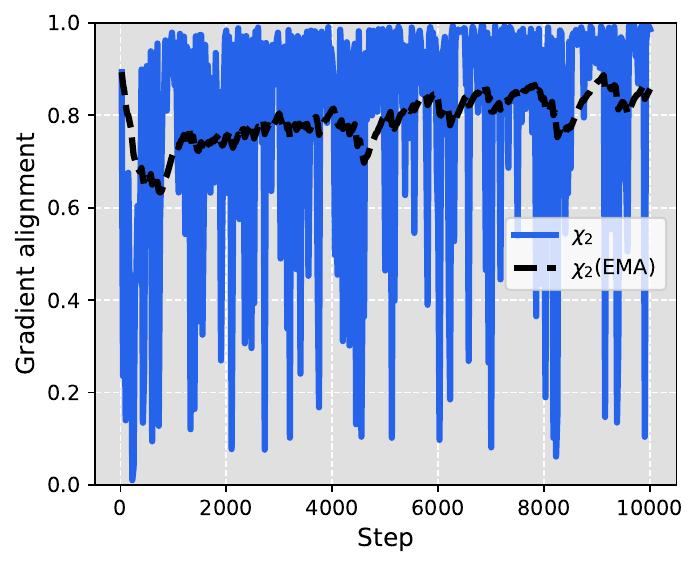} 
        \caption*{(c) Transformers on SST2-5k}
    \end{minipage}
    \caption{Alignment between the gradient and the dominant Hessian eigenspace over
training. Gradients become increasingly aligned with the dominant subspace,
consistent with the dominant-space concentration observed in centralized
training.}
    \label{app:fig:dom_alignment}
\end{figure}

\paragraph{Observation 1: Gradients along Local SGD iterates align with the dominant subspace.}
We first measure the alignment of the full-batch gradient with the Hessian's Dominant subspace along the Local SGD trajectory. Specifically, at each synchronization round, after parameter averaging, all workers share the averaged
parameter vector $\bar \theta$. We compute the full-batch gradient $\nabla f(\bar \theta)$ and Hessian $H(\bar \theta)$, form the dominant subspace $\mathcal{S}_C(\bar \theta)$, and report $\chi_C(\nabla f(\bar \theta);\bar \theta)$. We perform this analysis on three model--dataset pairs: a fully-connected
tanh network on MNIST, a ReLU CNN on CIFAR10, and a 2-layer Transformer on
SST2. 

Figure~\ref{app:fig:dom_alignment} plots $\chi_C(\nabla f(\bar \theta);\bar \theta)$ over training for all three
model--dataset pairs, and Figure~\ref{app:fig:eigenvalues} shows the corresponding evolution of the Hessian spectrum. Across all settings, the full-batch gradients remain strongly aligned with the dominant subspace, while the leading Hessian eigenvalues stay well separated from the rest of the spectrum. These observations
mirror the dominant-subspace alignment previously reported for single-worker SGD,
and motivate the next question: whether training can be carried out using only the
Dominant or Bulk components of the Local SGD update.

\begin{figure}[H]
    \centering
    \begin{minipage}[b]{0.325\textwidth}
        \centering
        \includegraphics[width=\textwidth]{figures/stage/FC-TANH/train_loss.pdf}
        \caption*{(a) FC on MNIST-5k}
    \end{minipage}
    \hfill
    \begin{minipage}[b]{0.325\textwidth}
        \centering
        \includegraphics[width=\textwidth]{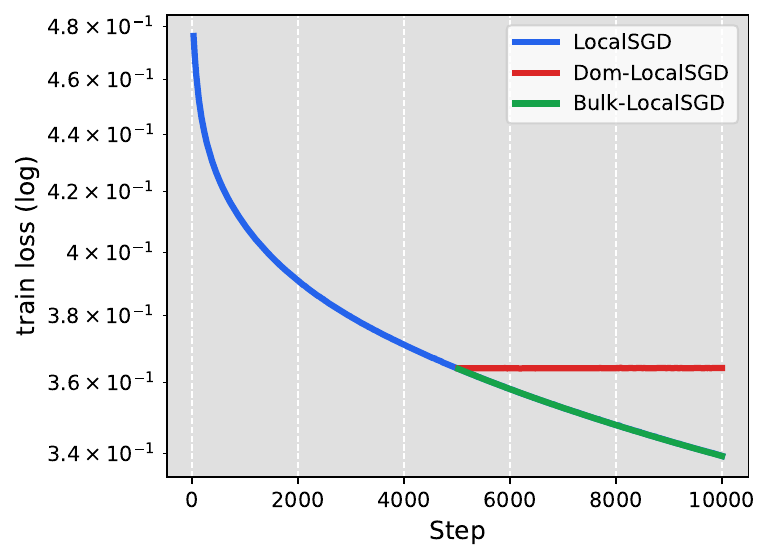}
        \caption*{(b) CNN on CIFAR10-5k}
    \end{minipage}
    \hfill
    \begin{minipage}[b]{0.325\textwidth}
        \centering
        \includegraphics[width=\textwidth]{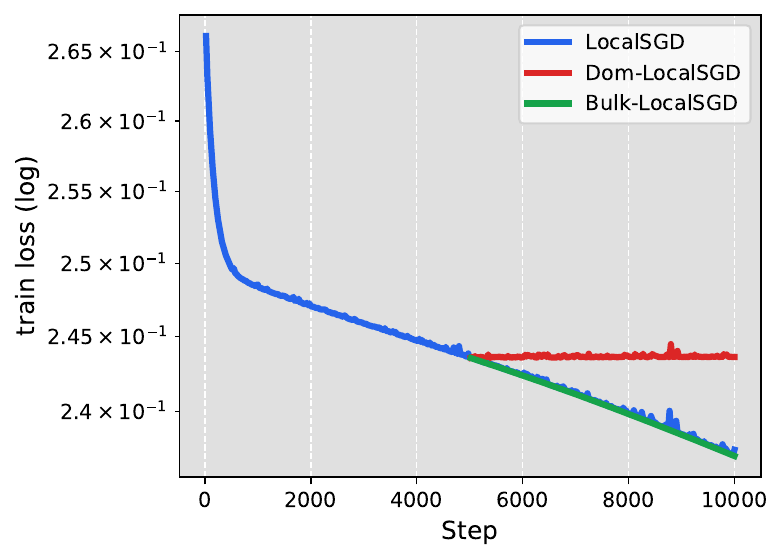} 
        \caption*{(c) Transformers on SST2-5k}
    \end{minipage}
    \caption{Training loss under Local SGD and updates restricted to the dominant or
bulk subspaces. Suppressing the bulk component slows or stalls progress, while
retaining the bulk component preserves useful descent, indicating that effective
learning relies strongly on the flatter bulk directions.}
    \label{app:fig:dom_bulk_training}
\end{figure}

\paragraph{Observation 2: Local SGD training cannot be performed in the Dominant Subspace.} 
Starting from checkpoints obtained during standard Local SGD training, we compare
three continuation runs: standard Local SGD, Dom-Local SGD, and Bulk-Local SGD.
Let $\bar \theta^c$ denote the synchronized average model at the beginning of
communication round $c$. All workers are initialized from this point,
$\theta_i^{c,0}=\bar \theta^c$, and perform $\tau$ local SGD steps to obtain
$\theta_i^{c,\tau}$. We define the average outer step over the
communication round as
\begin{equation}
    \bar p^c
    :=
    \frac{1}{M}\sum_{i=1}^M \theta_i^{c,\tau}
    -
    \bar \theta^c .
    \label{eq:avg_to_avg_displacement}
\end{equation}
Thus, standard Local SGD updates the synchronized model as
\begin{equation}
    \bar \theta^{c+1}
    =
    \bar \theta^c + \bar p^c .
\end{equation}

To isolate which part of this displacement is responsible for training progress,
we also consider projected continuation runs. Let
$\mathcal{S}_C(\bar \theta^c)$ denote the dominant subspace at $\bar \theta^c$, and let
$P_C^c$ be the orthogonal projection matrix onto this subspace. Dom-Local SGD keeps
only the component of the average outer step in the dominant
subspace,
\begin{equation}
    \bar \theta^{c+1}
    =
    \bar \theta^c + P_C^c \bar p^c ,
\end{equation}
whereas Bulk-Local SGD keeps only the complementary bulk component,
\begin{equation}
    \bar \theta^{c+1}
    =
    \bar \theta^c + (I-P_C^c)\bar p^c .
\end{equation}
In all three cases, the local trajectories are generated in the same way; the
only difference is whether the resulting communication-round displacement is
used directly, projected onto the dominant subspace, or projected onto the bulk
subspace.

Figure~\ref{app:fig:dom_bulk_training} shows that Dom-Local SGD fails to sustain
training, with the loss either stalling or diverging after the projection is
introduced. In contrast, Bulk-Local SGD closely matches standard Local SGD and
sometimes converges slightly faster. Thus, despite the strong dominant-subspace
alignment of full-batch gradients, the communication-round updates that sustain
training are primarily carried by the bulk component. This dominant--bulk
separation is consistent with the phenomenon previously observed in
single-worker SGD training.

\begin{figure}[H]
    \centering
    \begin{minipage}[b]{0.325\textwidth}
        \centering
        \includegraphics[width=\textwidth]{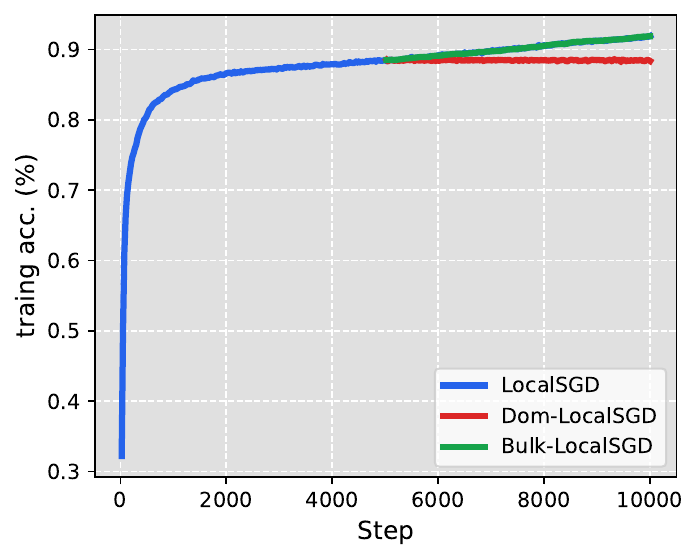}
        \caption*{(a) FC on MNIST-5k}
    \end{minipage}
    \hfill
    \begin{minipage}[b]{0.325\textwidth}
        \centering
        \includegraphics[width=\textwidth]{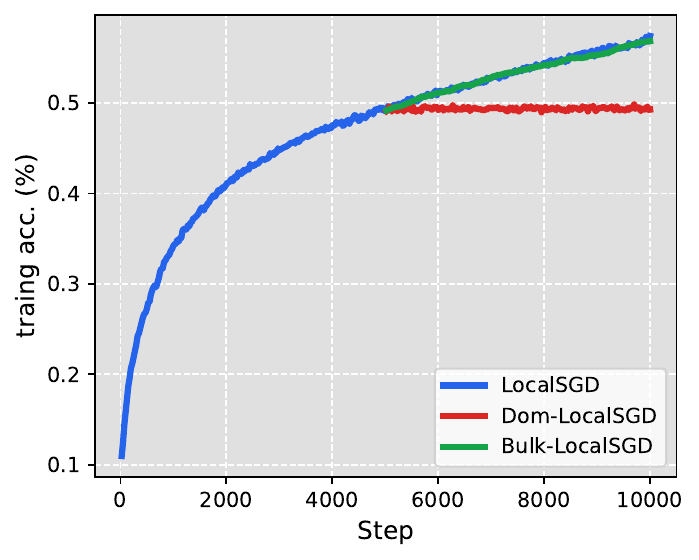}
        \caption*{(b) CNN on CIFAR10-5k}
    \end{minipage}
    \hfill
    \begin{minipage}[b]{0.325\textwidth}
        \centering
        \includegraphics[width=\textwidth]{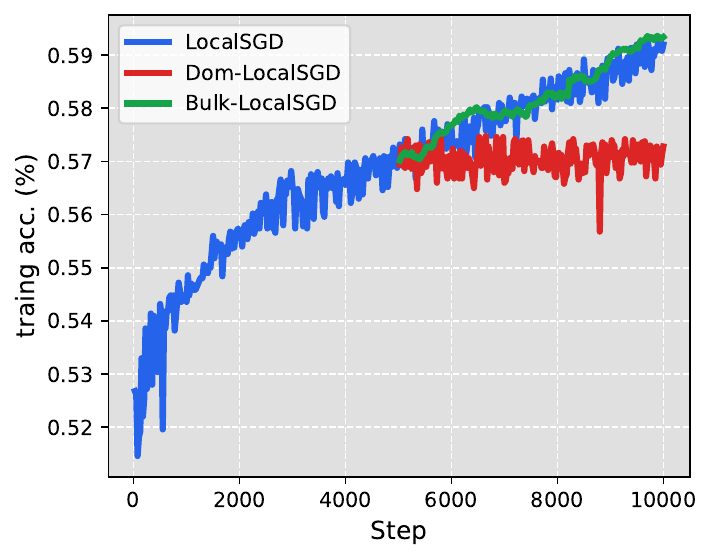} 
        \caption*{(c) Transformers on SST2-5k}
    \end{minipage}
    \caption{Training accuracy under Local SGD and updates restricted to the dominant or
bulk subspaces.}
    \label{app:fig:dom_bulk_train_acc}
\end{figure}

\begin{figure}[H]
    \centering
    \begin{minipage}[b]{0.325\textwidth}
        \centering
        \includegraphics[width=\textwidth]{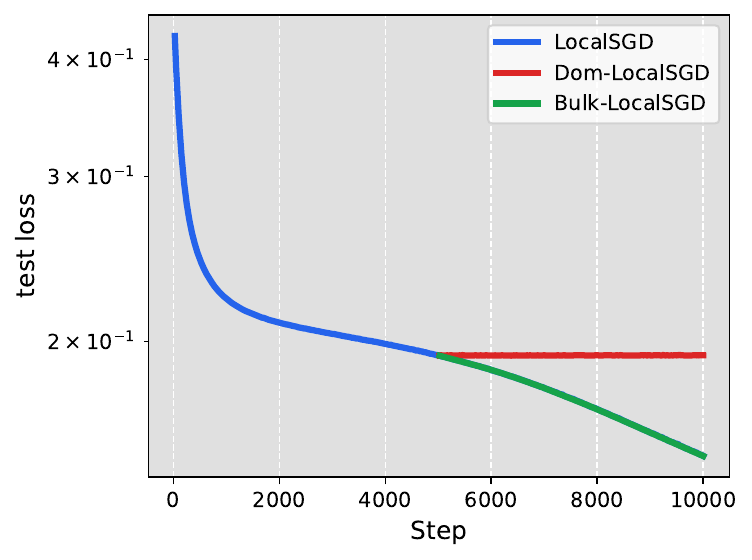}
        \caption*{(a) FC on MNIST-5k}
    \end{minipage}
    \hfill
    \begin{minipage}[b]{0.325\textwidth}
        \centering
        \includegraphics[width=\textwidth]{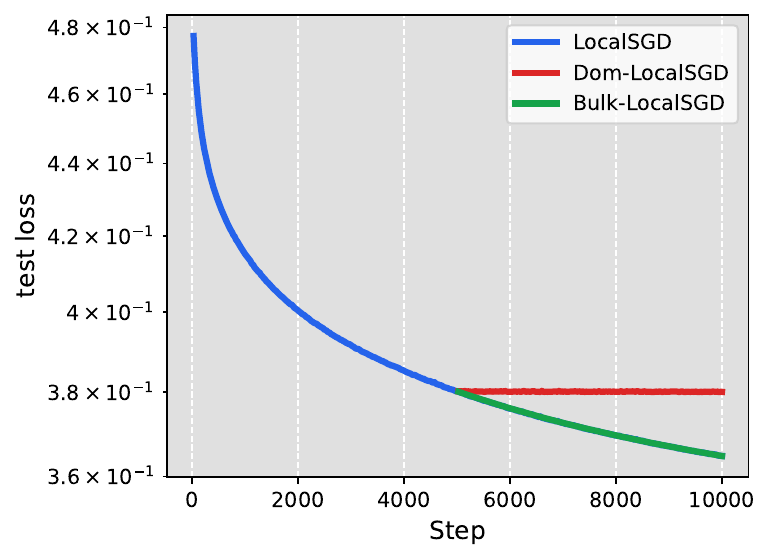}
        \caption*{(b) CNN on CIFAR10-5k}
    \end{minipage}
    \hfill
    \begin{minipage}[b]{0.325\textwidth}
        \centering
        \includegraphics[width=\textwidth]{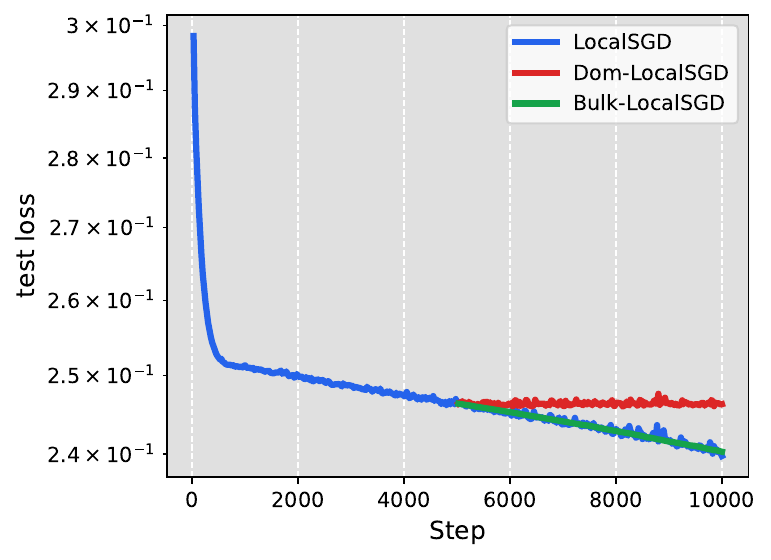} 
        \caption*{(c) Transformers on SST2-5k}
    \end{minipage}
    \caption{Test loss under Local SGD and updates restricted to the dominant or
bulk subspaces.}
    \label{app:fig:dom_bulk_test_loss}
\end{figure}

\begin{figure}[H]
    \centering
    \begin{minipage}[b]{0.325\textwidth}
        \centering
        \includegraphics[width=\textwidth]{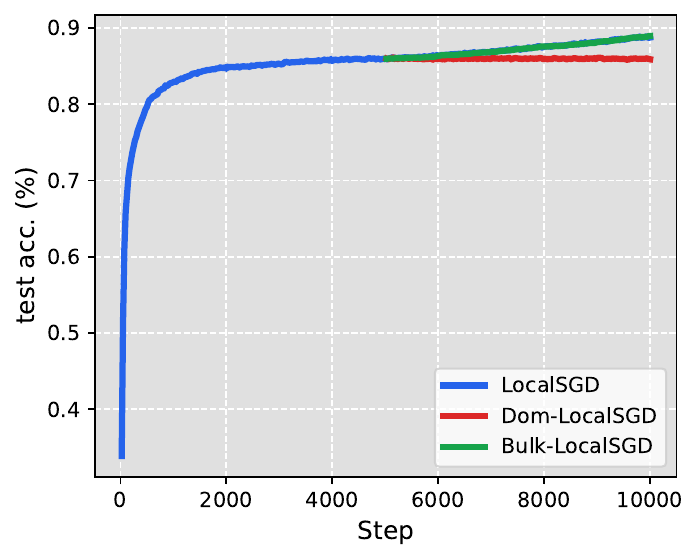}
        \caption*{(a) FC on MNIST-5k}
    \end{minipage}
    \hfill
    \begin{minipage}[b]{0.325\textwidth}
        \centering
        \includegraphics[width=\textwidth]{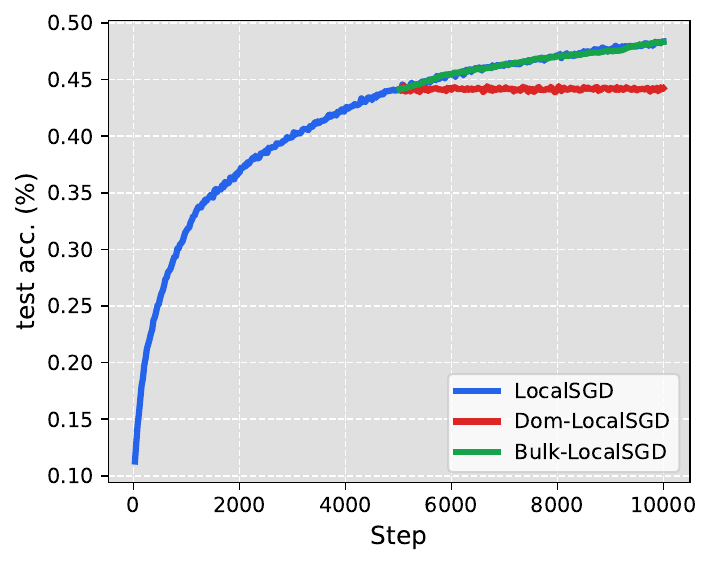}
        \caption*{(b) CNN on CIFAR10-5k}
    \end{minipage}
    \hfill
    \begin{minipage}[b]{0.325\textwidth}
        \centering
        \includegraphics[width=\textwidth]{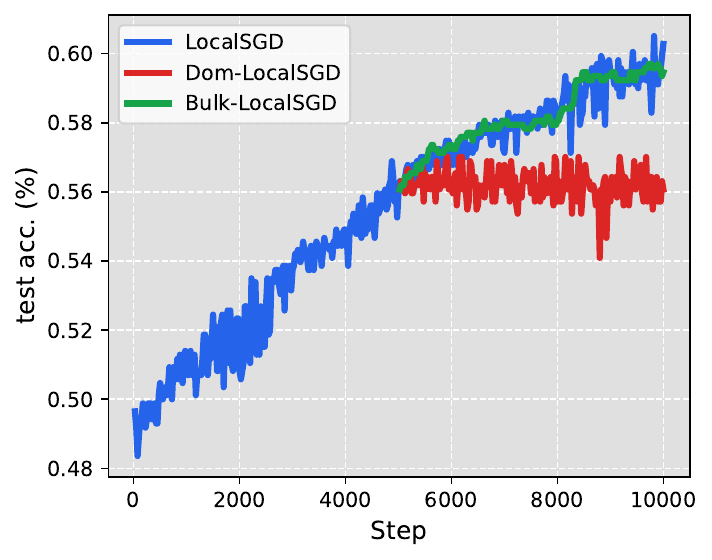} 
        \caption*{(c) Transformers on SST2-5k}
    \end{minipage}
    \caption{Test accuracy under Local SGD and updates restricted to the dominant or
bulk subspaces.}
    \label{app:fig:dom_bulk_test_acc}
\end{figure}

For completeness, we also share how train accuracy, test loss and test accuracy changes for different model--dataset pairs  in Figures \ref{app:fig:dom_bulk_train_acc}, \ref{app:fig:dom_bulk_test_loss} and \ref{app:fig:dom_bulk_test_acc} respectively.

\subsection{Dominant Subspace Estimation using Worker-Average Gaps}
\label{app:subsec:gap_subspace}

To form a proxy subspace for estimating the dominant Hessian subspace, we
maintain a FIFO buffer of worker--average gaps collected during standard
Local SGD training. At synchronization round \(c\), the worker--average gap is
\[
    \Delta_i^c
    :=
    \theta_i^{c,\tau}
    -
    \frac{1}{M}\sum_{j=1}^M \theta_j^{c,\tau}.
\]
Equivalently, since all workers start the round from the same synchronized
parameter vector \(\theta_i^{c,0}=\theta_j^{c,0}\), if
\(p_i^c:=\theta_i^{c,\tau}-\theta_i^{c,0}\) denotes the outer displacement or progress of worker \(i\), then
\[
    \Delta_i^c
    =
    p_i^c
    -
    \frac{1}{M}\sum_{j=1}^M p_j^c .
\]
Thus, the worker--average gap is exactly the deviation of each worker's
accumulated local update from the cross-worker average accumulated update.

Since \(\sum_{i=1}^M \Delta_i^c=0\), at most \(M-1\) linearly independent gaps
are obtained from each synchronization round. We insert these gaps into a FIFO
buffer
\[
    Z_c=[z_1,\ldots,z_B]\in\mathbb{R}^{D\times B},
\]
where each buffer entry \(z_j\) is a previously observed worker--average gap
\(\Delta_i^{c'}\), \(B\) is the buffer capacity, and \(D\) is the number of model
parameters. We form the proxy subspace using the full effective rank of the current buffer.
Concretely, we compute the Gram matrix
\[
    G_c := Z_c^\top Z_c,
\]
keep the eigenvectors whose eigenvalues are above a small relative threshold,
and write the retained eigendecomposition as
\[
    G_c V_c = V_c \Omega_c .
\]
We then construct an orthonormal basis for the retained column space of \(Z_c\)
as
\[
    Q_c := Z_c V_c \Omega_c^{-1/2}.
\]
The resulting subspace
\[
    \widehat{\mathcal S}_c := \operatorname{span}(Q_c)
\]
is our worker-gap proxy for directions of large recent worker disagreement. Its
effective rank is $B_c = \dim(\widehat{\mathcal S}_c)$, which equals the
number of retained eigenvalues of \(G_c\). In practice, this is typically the
buffer size \(B\), except when some eigenvalues of \(G_c\) fall below the
relative threshold \(10^{-8}\).

Algorithm~\ref{alg:gap_subspace_dom_removal} summarizes the construction of the
worker--average gap subspace and the dominant component removal metric $\rho$ used in our
experiments. At each synchronization round, we first compute the deviations of
the local worker parameters from their synchronized average and insert these
vectors into a FIFO buffer. We then compute an orthonormal basis for the span of
the buffered gaps using the Gram matrix $Z_c^\top Z_c$, which avoids explicitly
forming a large $D\times D$ covariance matrix. The resulting basis $Q_c$ defines
the proxy projector $P_{Q_c}=Q_cQ_c^\top$. Given the dominant Hessian basis
$U_C$, we measure how much of the true dominant component of the full-batch
gradient $g_c=\nabla f(\bar{\theta}^c)$ is removed by the gap-subspace filter
$I-P_{Q_c}$ using the ratio $\rho_c(g_c)$.

\begin{algorithm}[H]
\caption{Worker--Average Gap Subspace and Dominant Component Removal}
\label{alg:gap_subspace_dom_removal}
\begin{algorithmic}[1]
\REQUIRE Local worker parameters $\{\theta_i^{c,\tau}\}_{i=1}^M$ at synchronization round $c$,
buffer capacity $B$, relative threshold $\varepsilon$, dominant Hessian basis $U_C$,
full-batch gradient $g_c$
\STATE Compute the synchronized average
\(
    \bar{\theta}^{c}
    \leftarrow
    \frac{1}{M}
    \sum_{i=1}^{M}
    \theta_i^{c,\tau}
\)
\STATE Compute worker--average gaps
\(
    \Delta_i^c
    \leftarrow
    \theta_i^{c,\tau}
    -
    \bar{\theta}^{c},
    \qquad i=1,\ldots,M-1
\)
\hfill (only $M-1$ are linearly independent since $\sum_{i=1}^M \Delta_i^c=0$)
\STATE Insert $\{\Delta_i^c\}_{i=1}^{M-1}$ into the FIFO buffer and discard the oldest entries if the buffer size exceeds $B$
\STATE Let the current buffer be
\(
    Z_c = [z_1,\ldots,z_{B_c}] \in \mathbb{R}^{D\times B_c},
\)
where $B_c\leq B$ is the current number of stored gaps
\STATE Form the Gram matrix
\(
    G_c \leftarrow Z_c^\top Z_c
\)
\STATE Compute the eigendecomposition
\(
    G_c V_c = V_c \Omega_c
\)
\STATE Retain eigenpairs whose eigenvalues satisfy
\(
    \omega_r \geq \varepsilon \omega_{\max}(G_c)
\)
\STATE Let $V_c^{\mathrm{ret}}$ and $\Omega_c^{\mathrm{ret}}$ denote the retained eigenvectors and eigenvalues
\STATE Construct the orthonormal gap-subspace basis
\(
    Q_c
    \leftarrow
    Z_c V_c^{\mathrm{ret}}
    \left(\Omega_c^{\mathrm{ret}}\right)^{-1/2}
\)
\STATE Compute the dominant Hessian projector
\(
    P_C \leftarrow U_C U_C^\top
\)
\STATE Compute the worker-gap projector
\(
    P_{Q_c} \leftarrow Q_c Q_c^\top
\)
\STATE Compute the dominant component removal ratio
\(
    \rho_c(g_c)
    \leftarrow
    1 -
    \frac{
    \left\|P_C(I-P_{Q_c})g_c\right\|_2
    }{
    \left\|P_C g_c\right\|_2
    }
\)
\RETURN Gap-subspace basis $Q_c$ and dominant component removal ratio $\rho_c(g_c)$
\end{algorithmic}
\end{algorithm}

\begin{figure}[t]
    \centering
    \begin{minipage}[b]{0.325\textwidth}
        \centering
        \includegraphics[width=\textwidth]{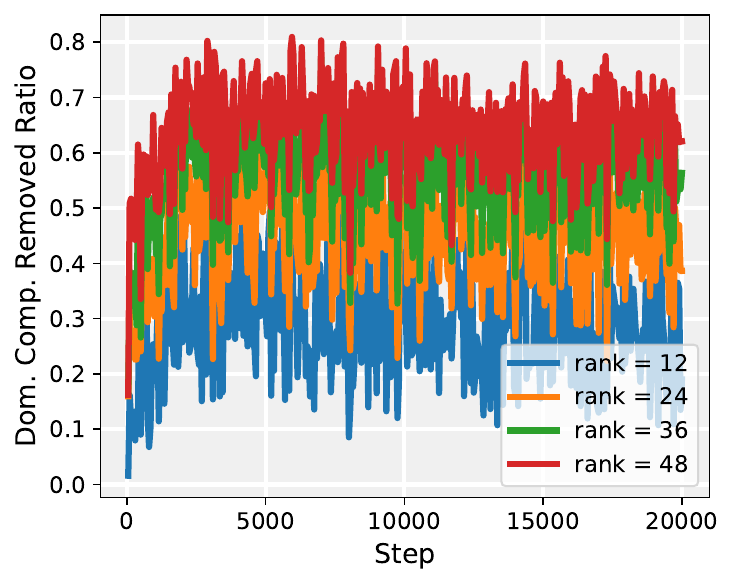}
        \caption*{(a) FC on MNIST-5k}
    \end{minipage}
    \hfill
    \begin{minipage}[b]{0.325\textwidth}
        \centering
        \includegraphics[width=\textwidth]{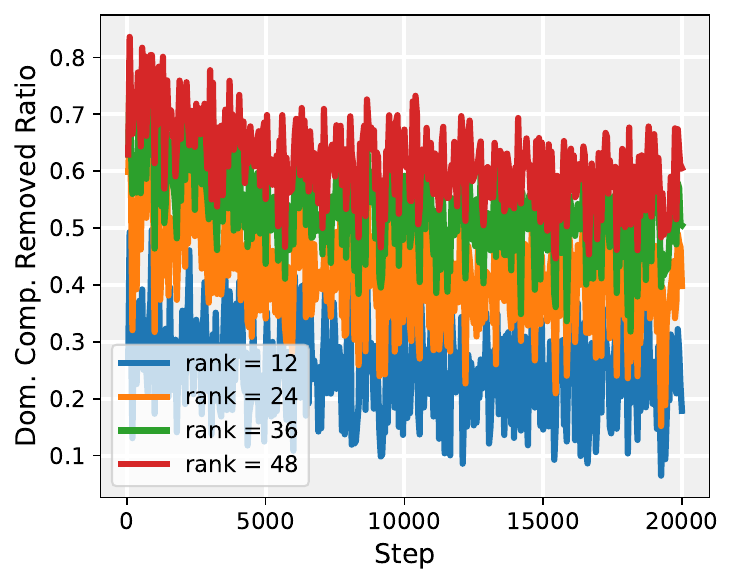}
        \caption*{(b) CNN on CIFAR10-5k}
    \end{minipage}
    \hfill
    \begin{minipage}[b]{0.325\textwidth}
        \centering
        \includegraphics[width=\textwidth]{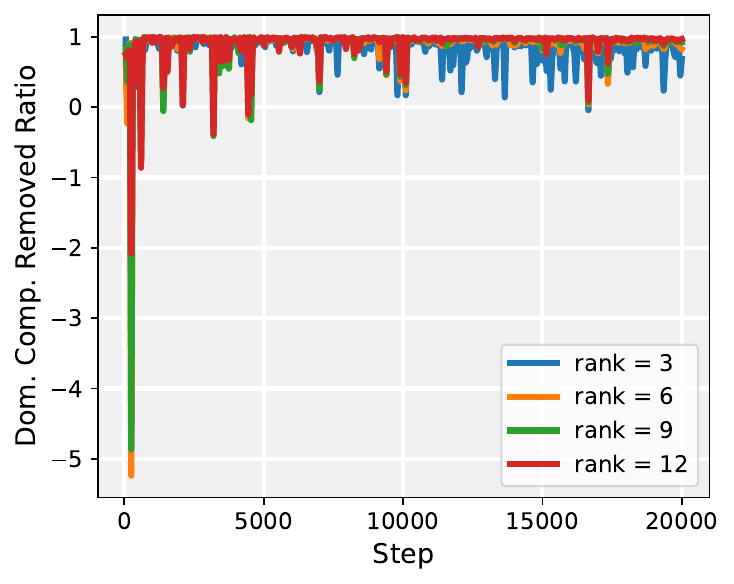} 
        \caption*{(c) Transformer on SST2-5k}
    \end{minipage}
    \vspace{-2mm}
    \caption{Raw dominant-component removal fraction achieved by worker-gap subspaces of
different buffer capacities. The worker-gap subspace captures a large portion of
the true dominant Hessian component across architectures, with coverage improving
as the buffer capacity increases.} 
\label{app:fig:raw_gap_subspace_results}
\vspace{-3mm}    
\end{figure}

In section~\ref{sec:main_empirical_results}, we share the EMA smoothed $\rho_c(\cdot)$ curves for clarity. For completeness, we also share the plots with the raw values in Figure~\ref{app:fig:raw_gap_subspace_results}.

\section{Theoretical Characterization of Worker-Average Gaps}
\label{app:full_theory}

Here, we provide full proofs of the theory presented in the main body of the paper. The  derivations of Lemma~\ref{lem:gap_recurrence}, Theorem~\ref{thm:gap_covariance} and Proposition~\ref{prop:noise_curvature} are presented in Subsections~\ref{app:sec:gap_curvature_model}, \ref{app:sec:gap_covariance_hessian} and \ref{app:sec:noise_curvature_coupling}

\subsection{Linearized worker--average gap dynamics}
\label{app:sec:gap_curvature_model}

We consider Local SGD training with \(M\) workers. Let
\(\theta_i^{c,t}\) denote the parameter vector of worker \(i\) after \(t\)
local steps in communication round \(c\), where \(t=0,\ldots,\tau\). At the
beginning of each round, all workers are synchronized:
\[
    \theta_i^{c,0}=\bar\theta^c,
    \qquad i=1,\ldots,M .
\]
Each worker performs local SGD updates using stochastic gradients of the form
\begin{equation}
    g_i^{c,t}
    =
    \nabla f(\theta_i^{c,t-1})
    +
    \epsilon_i^{c,t},
    \qquad
    \mathbb E[\epsilon_i^{c,t}\mid \theta_i^{c,t-1}]=0 .
    \label{eq:realistic_gradient_model}
\end{equation}
Thus,
\begin{equation}
    \theta_i^{c,t}
    =
    \theta_i^{c,t-1}
    -
    \eta
    \left[
    \nabla f(\theta_i^{c,t-1})
    +
    \epsilon_i^{c,t}
    \right].
    \label{eq:worker_local_update}
\end{equation}

To describe the relative motion of the workers within a communication round,
define the within-round average
\begin{equation}
    \bar\theta^{c,t}
    :=
    \frac{1}{M}\sum_{j=1}^{M}\theta_j^{c,t},
    \label{eq:within_round_average}
\end{equation}
and the worker deviation from this average
\begin{equation}
    d_i^{c,t}
    :=
    \theta_i^{c,t}-\bar\theta^{c,t}.
    \label{eq:within_round_deviation}
\end{equation}
By construction,
\[
    \frac{1}{M}\sum_{i=1}^M d_i^{c,t}=0,
    \qquad
    d_i^{c,0}=0 .
\]
After \(\tau\) local steps, the next synchronized model is
\[
    \bar\theta^{c+1}:=\bar\theta^{c,\tau},
\]
and the worker--average gap at communication round \(c+1\) is
\begin{equation}
    \Delta_i^{c+1}
    :=
    \theta_i^{c,\tau}-\bar\theta^{c+1}
    =
    d_i^{c,\tau}.
    \label{eq:final_worker_average_gap}
\end{equation}

We next linearize the population gradient around the within-round average
\(\bar\theta^{c,t-1}\). Define the local Hessian
\begin{equation}
    H_{c,t}
    :=
    \nabla^2 f(\bar\theta^{c,t-1}).
    \label{eq:local_hessian_time_varying}
\end{equation}
Then
\begin{equation}
    \nabla f(\theta_i^{c,t-1})
    =
    \nabla f(\bar\theta^{c,t-1})
    +
    H_{c,t}d_i^{c,t-1}
    +
    \mathcal R_i^{c,t},
    \label{eq:local_linearization_deviation}
\end{equation}
where \(\mathcal R_i^{c,t}\) denotes the higher-order Taylor remainder.

Averaging~\eqref{eq:worker_local_update} across workers gives
\begin{equation}
    \bar\theta^{c,t}
    =
    \bar\theta^{c,t-1}
    -
    \eta
    \left[
    \frac{1}{M}
    \sum_{j=1}^{M}
    \nabla f(\theta_j^{c,t-1})
    +
    \bar\epsilon^{c,t}
    \right],
    \qquad
    \bar\epsilon^{c,t}
    :=
    \frac{1}{M}\sum_{j=1}^{M}\epsilon_j^{c,t}.
    \label{eq:within_round_average_update}
\end{equation}
Define the centered worker noise
\begin{equation}
    \zeta_i^{c,t}
    :=
    \epsilon_i^{c,t}-\bar\epsilon^{c,t}.
    \label{eq:centered_worker_noise}
\end{equation}

Subtracting~\eqref{eq:within_round_average_update} from
\eqref{eq:worker_local_update} yields
\begin{align}
    d_i^{c,t}
    &=
    d_i^{c,t-1}
    -
    \eta
    \left(
    \nabla f(\theta_i^{c,t-1})
    -
    \frac{1}{M}\sum_{j=1}^{M}\nabla f(\theta_j^{c,t-1})
    \right)
    -
    \eta \zeta_i^{c,t}.
    \label{eq:deviation_exact_before_linearization}
\end{align}
Using~\eqref{eq:local_linearization_deviation} and
\(\frac{1}{M}\sum_{j=1}^{M}d_j^{c,t-1}=0\), we obtain
\begin{align}
    d_i^{c,t}
    =
    \left(I-\eta H_{c,t}\right)d_i^{c,t-1}
    -
    \eta \zeta_i^{c,t}
    -
    \eta
    \left(
    \mathcal R_i^{c,t}-\bar{\mathcal R}^{c,t}
    \right),
    \label{eq:deviation_linearized_with_remainder}
\end{align}
where
\[
    \bar{\mathcal R}^{c,t}
    :=
    \frac{1}{M}\sum_{j=1}^{M}\mathcal R_j^{c,t}.
\]
Ignoring the higher-order terms gives the approximate recurrence
\begin{equation}
    d_i^{c,t}
    \approx
    \left(I-\eta H_{c,t}\right)d_i^{c,t-1}
    -
    \eta \zeta_i^{c,t}.
    \label{eq:deviation_linearized_recurrence}
\end{equation}
This recurrence shows that the shared center gradient cancels across workers:
worker deviations are driven by centered stochastic-gradient noise, while the
local Hessian determines how existing deviations are propagated.

\subsection{Worker--average gap covariance as propagated stochastic noise}
\label{app:sec:gap_covariance_hessian}

We now specialize the recurrence~\eqref{eq:deviation_linearized_recurrence}
to obtain an explicit covariance expression. Assume that the Hessian does not
change substantially within one communication round, so that
\[
    H_{c,t}\approx H_c,
    \qquad t=1,\ldots,\tau .
\]
Then
\begin{equation}
    d_i^{c,t}
    \approx
    (I-\eta H_c)d_i^{c,t-1}
    -
    \eta \zeta_i^{c,t}.
    \label{eq:deviation_constant_hessian_recurrence}
\end{equation}
For a noise vector injected at local step \(s\), define the propagation matrix
from step \(s\) to the end of the local phase as
\begin{equation}
    A_{c,s}
    :=
    (I-\eta H_c)^{\tau-s}.
    \label{eq:constant_hessian_propagator}
\end{equation}
Unrolling~\eqref{eq:deviation_constant_hessian_recurrence} from
\(d_i^{c,0}=0\) gives
\begin{equation}
    \Delta_i^{c+1}
    =
    d_i^{c,\tau}
    \approx
    -\eta
    \sum_{s=1}^{\tau}
    A_{c,s}\zeta_i^{c,s}.
    \label{eq:gap_constant_hessian}
\end{equation}
This expression shows that worker--average gaps are not merely raw
stochastic-gradient noise. Rather, the noise injected at each local step is
propagated through the local optimization dynamics before synchronization. 
Assume that the stochastic-gradient noise is independent across workers and
local steps, with
\[
    \operatorname{Cov}(\epsilon_i^{c,t})=\Sigma_c.
\]
Since
\[
    \bar\epsilon^{c,t}
    =
    \frac{1}{M}\sum_{j=1}^{M}\epsilon_j^{c,t},
\]
we have
\[
    \zeta_i^{c,t}
    =
    \epsilon_i^{c,t}-\bar\epsilon^{c,t}
    =
    \left(1-\frac{1}{M}\right)\epsilon_i^{c,t}
    -
    \frac{1}{M}\sum_{j\neq i}\epsilon_j^{c,t}.
\]
Using independence across workers,
\begin{align}
    \operatorname{Cov}(\zeta_i^{c,t})
    &=
    \left(1-\frac{1}{M}\right)^2\Sigma_c
    +
    \frac{M-1}{M^2}\Sigma_c \notag \\
    &=
    \left(1-\frac{1}{M}\right)\Sigma_c .
    \label{eq:centered_noise_covariance}
\end{align}

Starting from~\eqref{eq:gap_constant_hessian}, and assuming
\(\mathbb E[\zeta_i^{c,s}]=0\), the gap covariance is
\begin{align}
    \operatorname{Cov}(\Delta_i^{c+1})
    &\approx
    \eta^2
    \sum_{s=1}^{\tau}
    \sum_{\ell=1}^{\tau}
    A_{c,s}
    \mathbb E
    \left[
    \zeta_i^{c,s}(\zeta_i^{c,\ell})^\top
    \right]
    A_{c,\ell}^\top .
    \label{eq:gap_cov_double_sum}
\end{align}
By independence across local steps, the cross terms vanish for \(s\neq \ell\),
and therefore
\begin{equation}
    \operatorname{Cov}(\Delta_i^{c+1})
    \approx
    \eta^2
    \sum_{s=1}^{\tau}
    A_{c,s}
    \operatorname{Cov}(\zeta_i^{c,s})
    A_{c,s}^\top .
    \label{eq:gap_cov_diagonal_terms}
\end{equation}
Substituting~\eqref{eq:centered_noise_covariance} gives
\begin{equation}
    \operatorname{Cov}(\Delta_i^{c+1})
    \approx
    \eta^2
    \left(1-\frac{1}{M}\right)
    \sum_{s=1}^{\tau}
    A_{c,s}\Sigma_cA_{c,s}^\top .
    \label{eq:gap_cov_with_propagators}
\end{equation}
Using \(A_{c,s}=(I-\eta H_c)^{\tau-s}\), the symmetry of \(H_c\), and
reindexing with \(q=\tau-s\), we obtain
\begin{equation}
    \boxed{
    \operatorname{Cov}(\Delta_i^{c+1})
    \approx
    \eta^2
    \left(1-\frac{1}{M}\right)
    \sum_{q=0}^{\tau-1}
    (I-\eta H_c)^q
    \Sigma_c
    (I-\eta H_c)^q
    }.
    \label{app:eq:gap_covariance_curvature_derivation}
\end{equation}
Thus, the worker-gap covariance is shaped jointly by the stochastic-gradient
noise covariance \(\Sigma_c\) and the local curvature \(H_c\).

\subsection{Directional gap variance under noise–curvature coupling}
\label{app:sec:noise_curvature_coupling}

Equation~\eqref{app:eq:gap_covariance_curvature_derivation} shows that the
worker-gap covariance is determined by the interaction between the local
curvature \(H_c\) and the stochastic-gradient noise covariance \(\Sigma_c\).
To interpret this expression directionally, let
\[
    H_c = U_c\Lambda_c U_c^\top,
    \qquad
    \Lambda_c=\operatorname{diag}(\lambda_{1,c},\ldots,\lambda_{D,c}),
\]
where \(u_{r,c}\), the \(r\)-th column of \(U_c\), is the Hessian eigenvector
associated with eigenvalue \(\lambda_{r,c}\). Assume that the noise covariance
is approximately diagonal in this Hessian eigenbasis:
\[
    U_c^\top \Sigma_c U_c
    \approx
    \operatorname{diag}
    \left(
    \sigma_{1,c}^2,\ldots,\sigma_{D,c}^2
    \right).
\]
Equivalently, \(\sigma_{r,c}^2\) denotes the stochastic-gradient noise variance
along Hessian eigendirection \(u_{r,c}\).

Projecting~\eqref{app:eq:gap_covariance_curvature_derivation} onto \(u_{r,c}\)
then gives
\begin{equation}
    \operatorname{Var}
    \left(
    \langle \Delta_i^{c+1},u_{r,c}\rangle
    \right)
    \approx
    \eta^2
    \left(1-\frac{1}{M}\right)
    \sigma_{r,c}^2
    \sum_{q=0}^{\tau-1}
    (1-\eta\lambda_{r,c})^{2q}.
    \label{app:eq:gap_variance_eigendirection}
\end{equation}
Define
\[
    \psi_\tau(a)
    :=
    \sum_{q=0}^{\tau-1}(1-a)^{2q}.
\]
Then~\eqref{app:eq:gap_variance_eigendirection} can be written as
\begin{equation}
    \operatorname{Var}
    \left(
    \langle \Delta_i^{c+1},u_{r,c}\rangle
    \right)
    \approx
    \eta^2
    \left(1-\frac{1}{M}\right)
    \sigma_{r,c}^2
    \psi_\tau(\eta\lambda_{r,c}).
    \label{app:eq:gap_variance_filter}
\end{equation}
Thus, the worker deviation along a Hessian eigendirection is controlled by two
factors: the stochastic-gradient noise strength \(\sigma_{r,c}^2\) along that
direction, and the curvature-dependent local-dynamics factor
\(\psi_\tau(\eta\lambda_{r,c})\).

Recent work on SGD noise and loss curvature suggests that the first factor is
itself coupled to curvature. In particular, when the SGD noise covariance is
expressed in the Hessian eigenbasis, its diagonal entries approximately obey
\[
    \sigma_{r,c}^2
    \propto
    \lambda_{r,c}^{\gamma},
    \qquad
    1\leq \gamma\leq 2,
\]
with superlinear scaling \(\gamma>1\) observed in cross-entropy
classification settings, while mean-squared error yields \(\gamma\approx 1\).
Under this empirically supported noise--curvature relation,
\eqref{app:eq:gap_variance_filter} gives
\begin{equation}
    \operatorname{Var}
    \left(
    \langle \Delta_i^{c+1},u_{r,c}\rangle
    \right)
    \propto
    \lambda_{r,c}^{\gamma}
    \psi_\tau(\eta\lambda_{r,c}).
    \label{app:eq:gap_variance_noise_curvature_scaling}
\end{equation}
This expression directly links worker disagreement to Hessian curvature:
directions with larger Hessian eigenvalues induce larger stochastic-gradient
noise, and therefore larger worker-average deviations, up to the modulation
introduced by \(\psi_\tau(\eta\lambda_{r,c})\). Hence, under the observed
noise--curvature coupling, worker-average gaps are expected to concentrate
along high-curvature Hessian directions. This provides a theoretical
motivation for using the span of worker-average gaps as a data-driven
estimator of the dominant curvature subspace.

\section{Additional Results}
\label{app:sec:additional_results}
In this section, we provide additional experimental results.

\subsection{Effect of the Communication Period on the Gap-Based Subspace}
\label{app:subsec:tau_ablation}

We investigate how the quality of the worker--average gap subspace changes with
the communication period $\tau$. To this end, we repeat the experimental setup
described in Section~\ref{sec:main_empirical_results} that uses three
model--dataset pairs: a tanh fully connected network on MNIST-5k, a ReLU CNN on
CIFAR-10-5k, and a 2-layer Transformer on SST-2-5k. As in the main experiments,
Local SGD is performed with $M=4$ workers.

For each setting, we construct the worker--average gap subspace using a FIFO
buffer of recent gaps and evaluate its ability to capture the dominant Hessian
subspace. We set the number of dominant Hessian directions to
$C=10$ for MNIST-5k and CIFAR-10-5k, and to
$C=2$ for SST-2-5k. For MNIST-5k and CIFAR-10-5k, we sweep buffer
capacities $B\in\{12,24,48\}$, while for SST-2-5k, we sweep
$B\in\{3,6,12\}$. We measure the effectiveness of the gap-induced subspace using
the dominant component removal ratio $\rho_c(v)$ defined in
Equation~\ref{eq:dom_mass_removed_frac}, evaluated throughout training. We
repeat the analysis for communication periods
$\tau\in\{2,5,10\}$. This way we can assess whether more frequent synchronization, which
produces shorter and more locally linear worker trajectories, leads to a more
accurate gap-based estimator of the dominant Hessian subspace. 

\begin{figure}[H]
    \centering
    \includegraphics[width=0.7\linewidth]{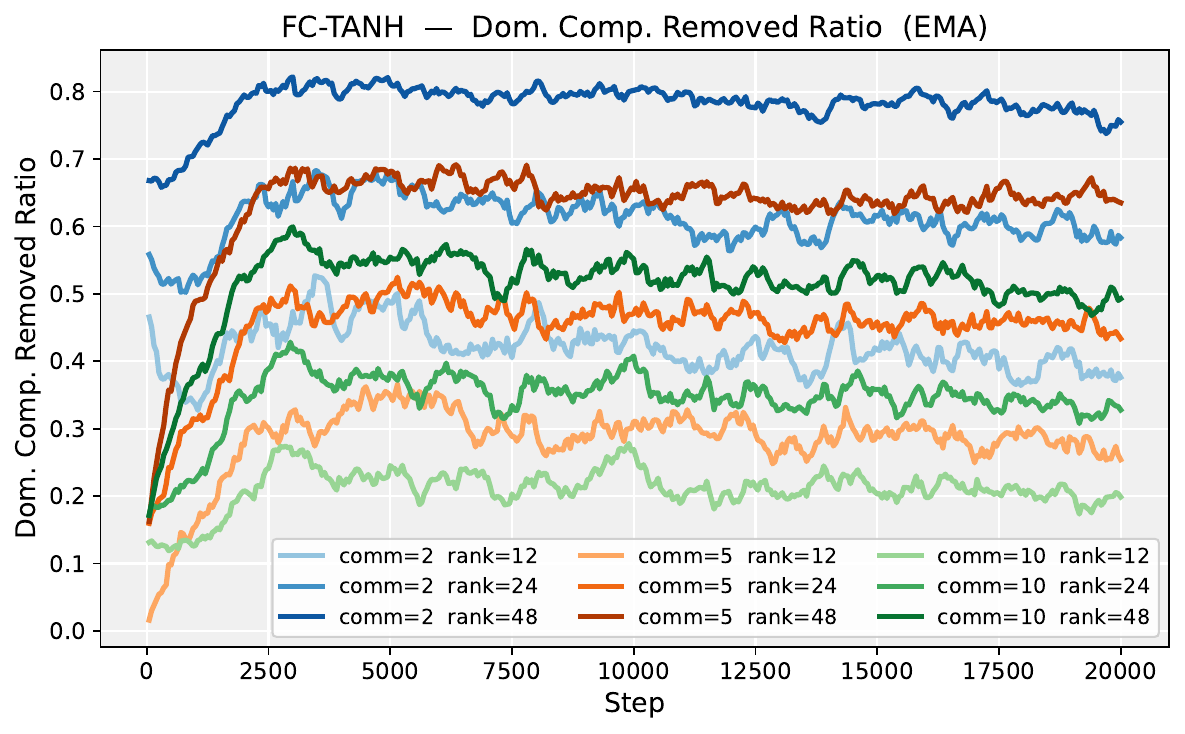}
    \caption{Communication period ablation for FC trained on MNIST-5k}
    \label{app:fig:comm_period_fc}
\end{figure}

\begin{figure}[H]
    \centering
    \includegraphics[width=0.7\linewidth]{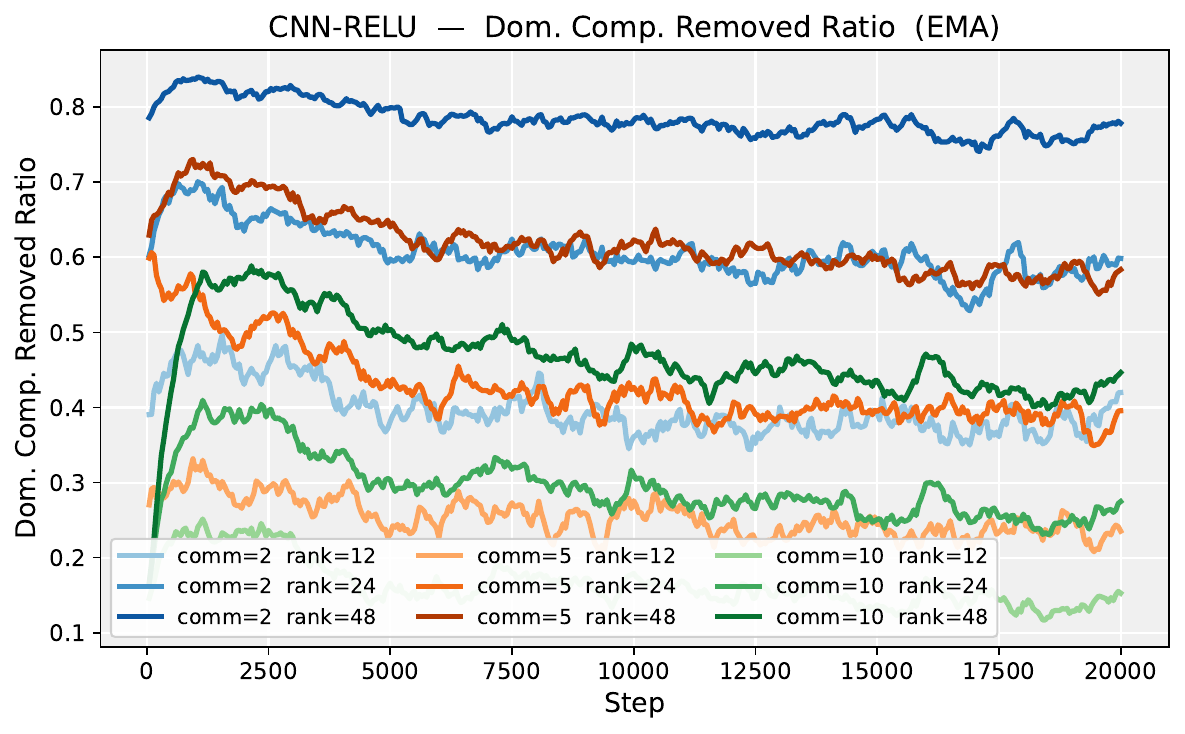}
    \caption{Communication period ablation for CNN trained on CIFAR10-5k}
    \label{app:fig:comm_period_cnn}
\end{figure}

\begin{figure}[H]
    \centering
    \includegraphics[width=0.7\linewidth]{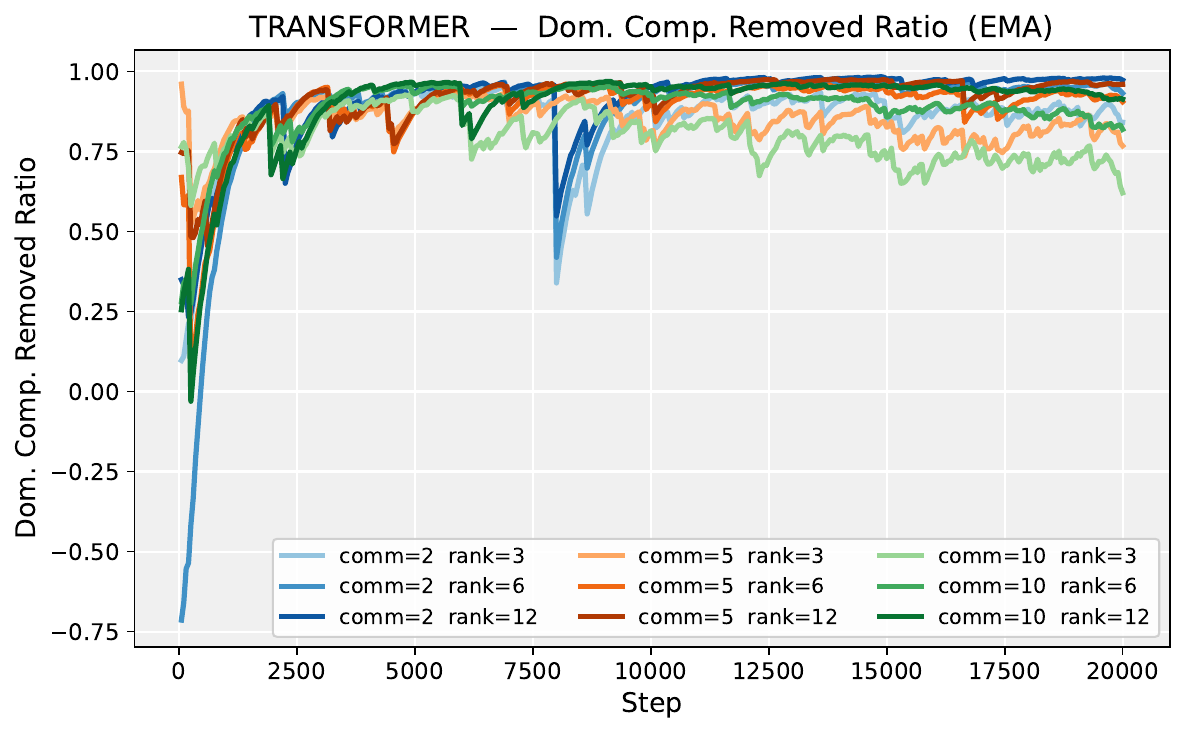}
    \caption{Communication period ablation for Transformer trained on SST2-5k}
    \label{app:fig:comm_period_transformer}
\end{figure}

We report the results in Figures~\ref{app:fig:comm_period_fc},
\ref{app:fig:comm_period_cnn}, and
\ref{app:fig:comm_period_transformer}. In each plot, curves corresponding to
the same communication period are grouped using the same color family, with
different shades indicating different buffer ranks. Different communication
periods are shown using distinct colors.

The effect of the communication period is consistent across FC-Tanh and
CNN-ReLU. Smaller communication periods yield substantially higher dominant component
removal. In particular, $\tau=2$ consistently achieves the largest $\rho_c(v)$,
followed by $\tau=5$, while $\tau=10$ gives the weakest removal. A striking
observation is that the $\tau=2$, $B=24$ configuration is approximately as
effective as the $\tau=5$, $B=48$ configuration in both settings. This suggests
that more frequent synchronization can compensate for a smaller gap buffer,
because shorter local trajectories produce worker--average gaps that more
faithfully reflect the local curvature--noise geometry around the synchronized
model. This trend is also consistent with the constant-Hessian approximation
used in the gap covariance analysis in
Section~\ref{app:sec:gap_covariance_hessian}: for smaller $\tau$, the Hessian is
more likely to remain approximately stable over the local trajectory, whereas
for larger $\tau$, nonlinear trajectory drift and Hessian variation can make the
collected gaps a less accurate proxy for the dominant Hessian subspace at the
synchronization point.

The Transformer on SST-2-5k shows a weaker dependence on the communication
period. Across most buffer ranks and communication periods, the dominant component
removal quickly becomes large and remains close to one throughout training. This
is likely because the dominant subspace dimension in this setting is only
$C=2$, making it easier for even a small worker-gap buffer to
capture the relevant directions. Overall, these results support the view that
worker--average gaps provide an effective Hessian-free estimator of the dominant
subspace, with the strongest and most consistent performance obtained when
synchronization is frequent and the buffer has sufficient rank.

\subsection{Can We Accelerate Optimization by Filtering Out the Dominant Component?}
\label{app:sec:gap_based_filtering}

In this section, we operationalize the use of worker--average gap-based
subspaces and ask whether training can be accelerated by suppressing the
estimated dominant directions and amplifying the orthogonal component. Our focus
here is not on generalization or test performance, but rather on training.
We evaluate whether the proposed filtering changes the rate at which the
training loss decreases by suppressing the estimated optimization steps 
along the high-curvature directions.

Inspired by the update modification in \cite{zhou-etal-2025-bsfa}, we propose the following:
Using the proxy subspace construction from Appendix~\ref{app:subsec:gap_subspace},
we modify the Local SGD synchronization step as follows. At communication round
\(c\), let
\[
    \bar p^c := \frac{1}{M}\sum_{i=1}^M p_i^c
\]
denote the standard Local SGD average outer displacement. Given the current
gap-based basis \(Q_c\), whose columns span the worker-gap proxy subspace
\(\widehat{\mathcal S}_c\), we decompose \(\bar p^c\) into its component inside
the estimated dominant subspace and its orthogonal complement:
\[
    p_{\mathrm{dom}}^c := Q_c Q_c^\top \bar p^c,
    \qquad
    p_{\mathrm{bulk}}^c := \bar p^c - p_{\mathrm{dom}}^c .
\]
We then replace the standard Local SGD synchronization update \(\bar p^c\) with
the filtered update
\[
    p_{\mathrm{filt}}^c
    :=
    \alpha p_{\mathrm{dom}}^c
    +
    \gamma p_{\mathrm{bulk}}^c ,
\]
and synchronize all workers to
\[
    \theta^{c+1,0}
    =
    \theta^{c,0}
    +
    p_{\mathrm{filt}}^c .
\]
Here, \(\alpha\) controls how strongly we retain the gap-estimated dominant
component, while \(\gamma\) controls the scaling of the orthogonal component.
Thus, choosing \(\alpha<1\) suppresses directions in
\(\widehat{\mathcal S}_c\), and choosing \(\gamma>1\) amplifies the component
orthogonal to the worker-gap subspace. The special case
\(\alpha=\gamma=1\) recovers standard Local SGD.

Similar to the experiment setup before, we train FC tanh on MNIST-5k, CNN ReLU on
CIFAR-10-5k, and a 2-layer Transformer on SST-2-5k.
In all settings, we perform training with Local SGD using $M=4$ workers and communication
period $\tau=5$. Each stochastic mini-batch has size 50, and training is run for
10,000 local steps, corresponding to 2,000 communication rounds. Since our
goal is to evaluate whether the Local SGD update modification leveraging gap-subspaces
can improve optimization, we first perform a careful learning-rate sweep for
each setting. We select the learning rate $\eta$ that achieves the lowest final
training loss. The sweep values are listed below, with the selected learning
rate shown in bold.

\begin{itemize}
    \item \textbf{FC tanh:} $[0.001, 0.005, 0.01, 0.05, 0.1, 0.2, \mathbf{0.3}, 0.4, 0.5]$
    \item \textbf{CNN ReLU:} $[0.001, 0.005, 0.01, \mathbf{0.02}, 0.03, 0.04, 0.05]$
    \item \textbf{Transformer:} $[0.01, 0.02, 0.03, 0.04, 0.06, 0.08, \mathbf{0.1}, 0.2]$
\end{itemize}

Next, we examine how the training loss evolves under two variants of the
filtered Local SGD update: (i) fixing \(\gamma=1.0\) and sweeping \(\alpha\),
and (ii) fixing \(\alpha=1.0\) and sweeping \(\gamma\). For the FC-TANH and
CNN-RELU experiments, we set the FIFO buffer capacity of the gap-based proxy
subspace to \(B=24\). For the Transformer experiments, we set \(B=6\). Recall
that the effective rank of the proxy subspace is determined by the number of
retained directions in the buffer span.

\begin{figure}[H]
    \centering
    \begin{minipage}[b]{0.49\textwidth}
        \centering
        \includegraphics[width=\textwidth]{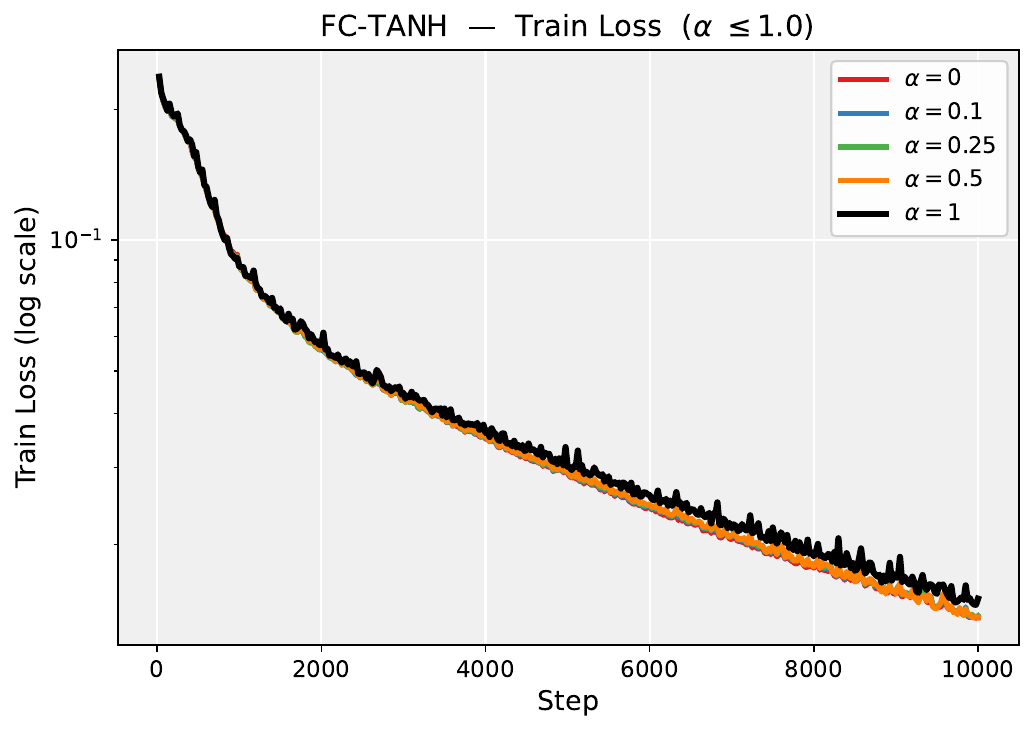}
    \end{minipage}
    \hfill
    \begin{minipage}[b]{0.49\textwidth}
        \centering
        \includegraphics[width=\textwidth]{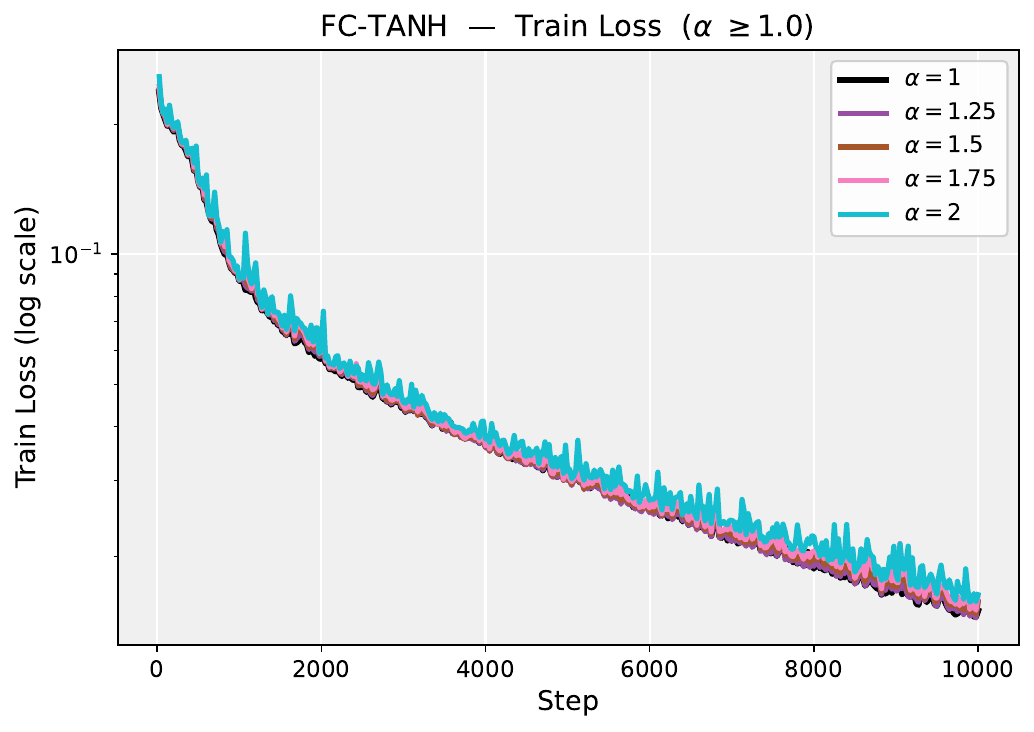}
    \end{minipage}
    \caption{FC TANH model trained on MNIST-5k. We fix $\gamma=1.0$ and vary $\alpha$.} 
\label{app:fig:fc_gamma_fix_alpha_sweep}
\end{figure}

\begin{figure}[H]
    \centering
    \begin{minipage}[b]{0.49\textwidth}
        \centering
        \includegraphics[width=\textwidth]{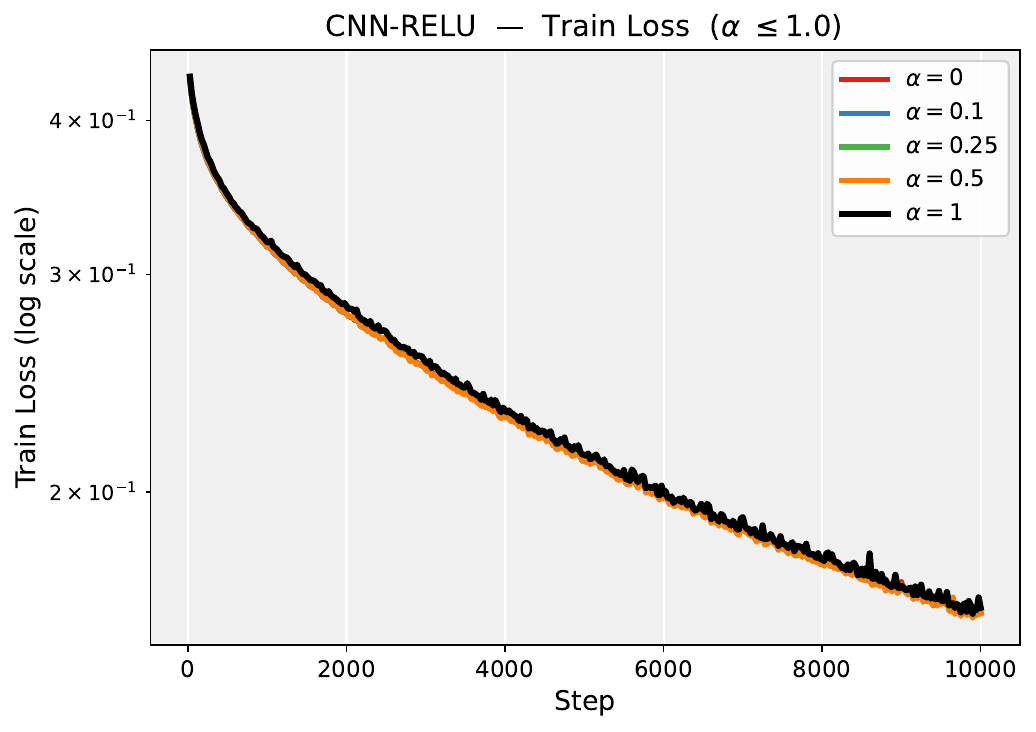}
    \end{minipage}
    \hfill
    \begin{minipage}[b]{0.49\textwidth}
        \centering
        \includegraphics[width=\textwidth]{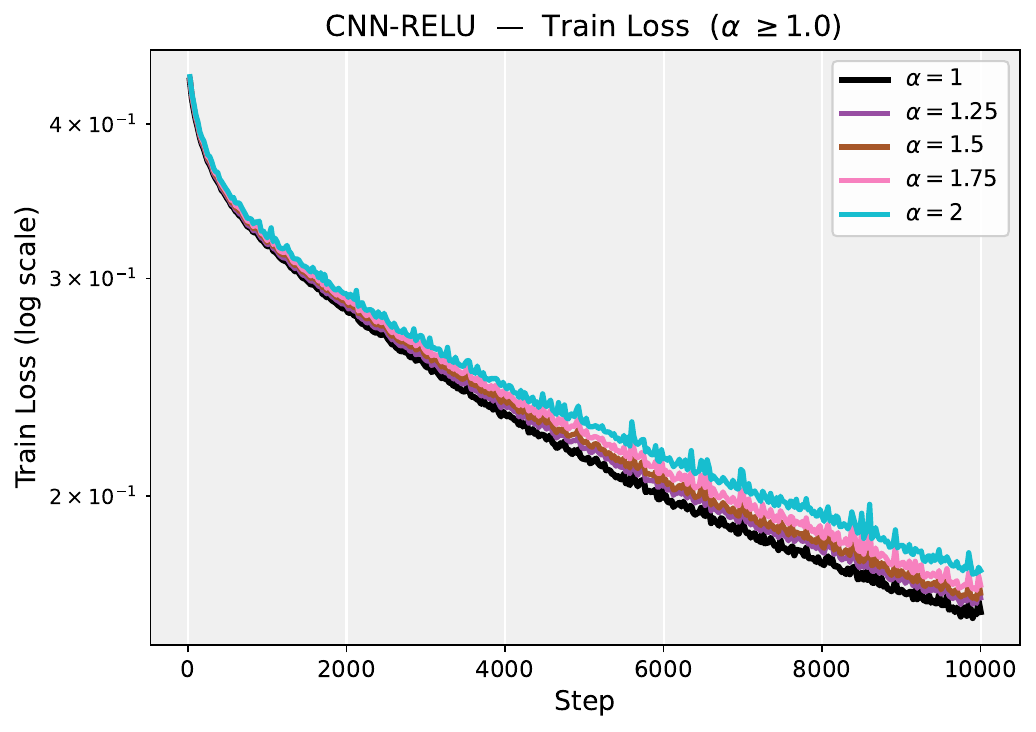}
    \end{minipage}
    \caption{CNN ReLU model trained on CIFAR10-5k. We fix $\gamma=1.0$ and vary $\alpha$.} 
\label{app:fig:cnn_gamma_fix_alpha_sweep}
\end{figure}

\begin{figure}[H]
    \centering
    \begin{minipage}[b]{0.49\textwidth}
        \centering
        \includegraphics[width=\textwidth]{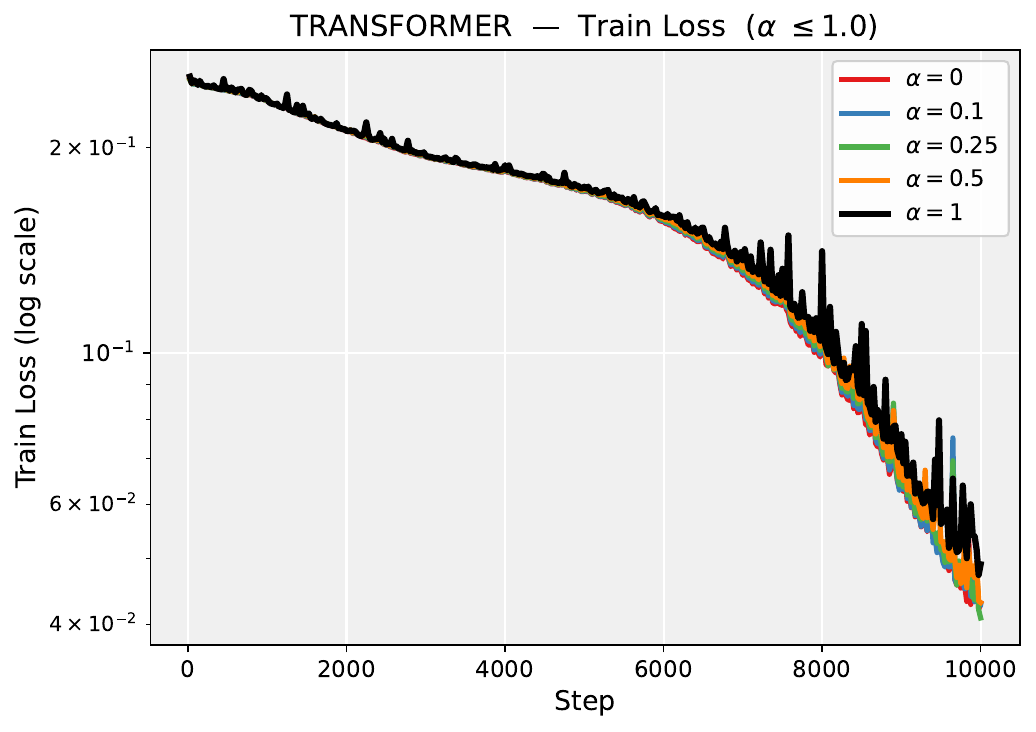}
    \end{minipage}
    \hfill
    \begin{minipage}[b]{0.49\textwidth}
        \centering
        \includegraphics[width=\textwidth]{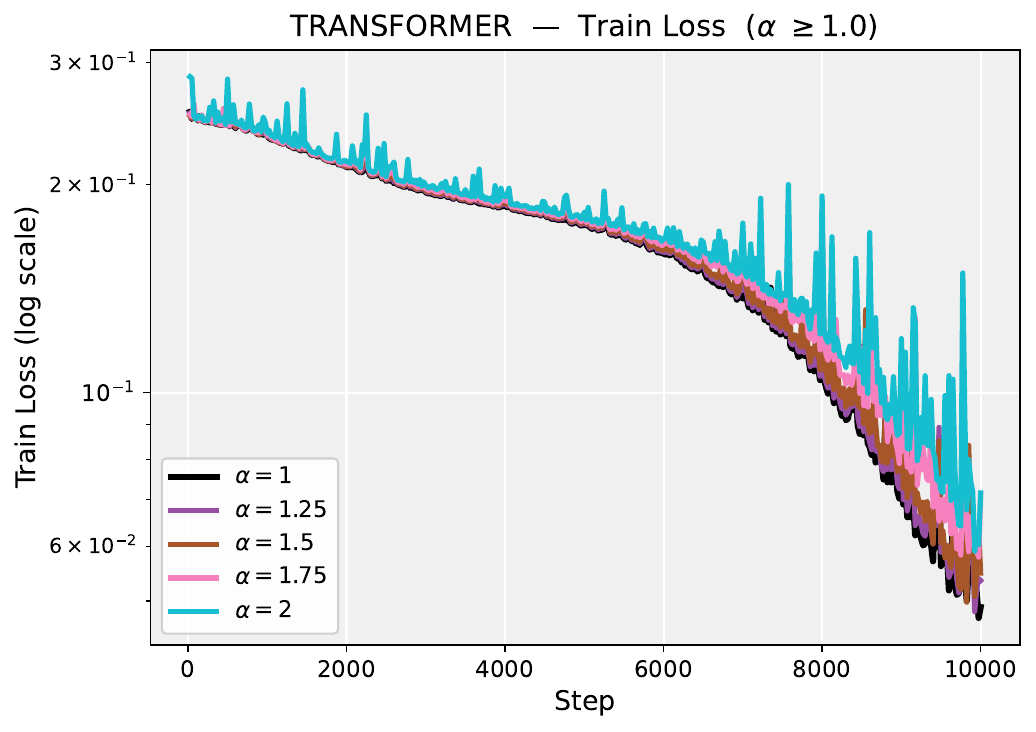}
    \end{minipage}
    \caption{Transformer model trained on SST-5k. We fix $\gamma=1.0$ and vary $\alpha$.} 
\label{app:fig:transformer_gamma_fix_alpha_sweep}
\end{figure}

\paragraph{i) Fixing \(\gamma=1.0\) and varying \(\alpha\).}
In this setting, we fix \(\gamma=1.0\) and sweep
\[
    \alpha \in
    \{0.0, 0.1, 0.25, 0.5, 1.0, 1.25, 1.5, 1.75, 2.0\}.
\]
When \(\alpha=1.0\), the filtered update reduces to standard Local SGD, and we
mark this baseline curve in black. We report the results in
Figures~\ref{app:fig:fc_gamma_fix_alpha_sweep},
\ref{app:fig:cnn_gamma_fix_alpha_sweep}, and
\ref{app:fig:transformer_gamma_fix_alpha_sweep}. In each figure, the left panel
shows the training loss curves for \(\alpha \leq 1.0\), corresponding to
suppression of the gap-estimated dominant component, while the right panel shows
the curves for \(\alpha \geq 1.0\), corresponding to amplification of this
component.

As can be seen from Figures~\ref{app:fig:fc_gamma_fix_alpha_sweep},
\ref{app:fig:cnn_gamma_fix_alpha_sweep}, and
\ref{app:fig:transformer_gamma_fix_alpha_sweep}, across all three settings, suppressing the gap-estimated dominant component
tends to improve the optimization trajectory relative to standard Local SGD.
When \(\gamma=1.0\) and \(\alpha<1.0\), the training loss decreases faster and
often reaches a lower final value than the \(\alpha=1.0\) baseline. This effect
is especially visible in the Transformer experiment, where reducing the weight
on \(p_{\mathrm{dom}}^c\) leads to a noticeably faster late-stage decrease in
training loss. Conversely, amplifying the same component by setting
\(\alpha>1.0\) slows down optimization and leads to worse final training loss.
The degradation becomes more pronounced as \(\alpha\) increases, particularly
for the Transformer model, where large \(\alpha\) values also introduce visibly
higher instability. These results support the view that the worker-gap subspace
captures directions that are detrimental to fast optimization, and that
suppressing the corresponding component of the Local SGD outer update can
accelerate training.

\begin{figure}[H]
    \centering
    \begin{minipage}[b]{0.49\textwidth}
        \centering
        \includegraphics[width=\textwidth]{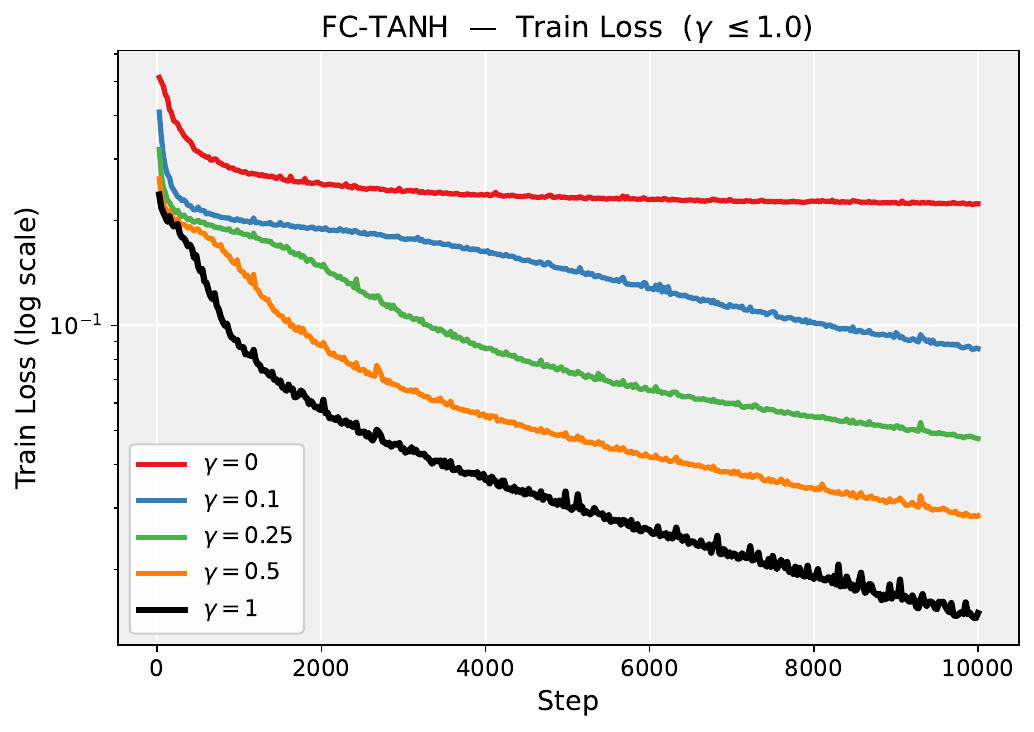}
    \end{minipage}
    \hfill
    \begin{minipage}[b]{0.49\textwidth}
        \centering
        \includegraphics[width=\textwidth]{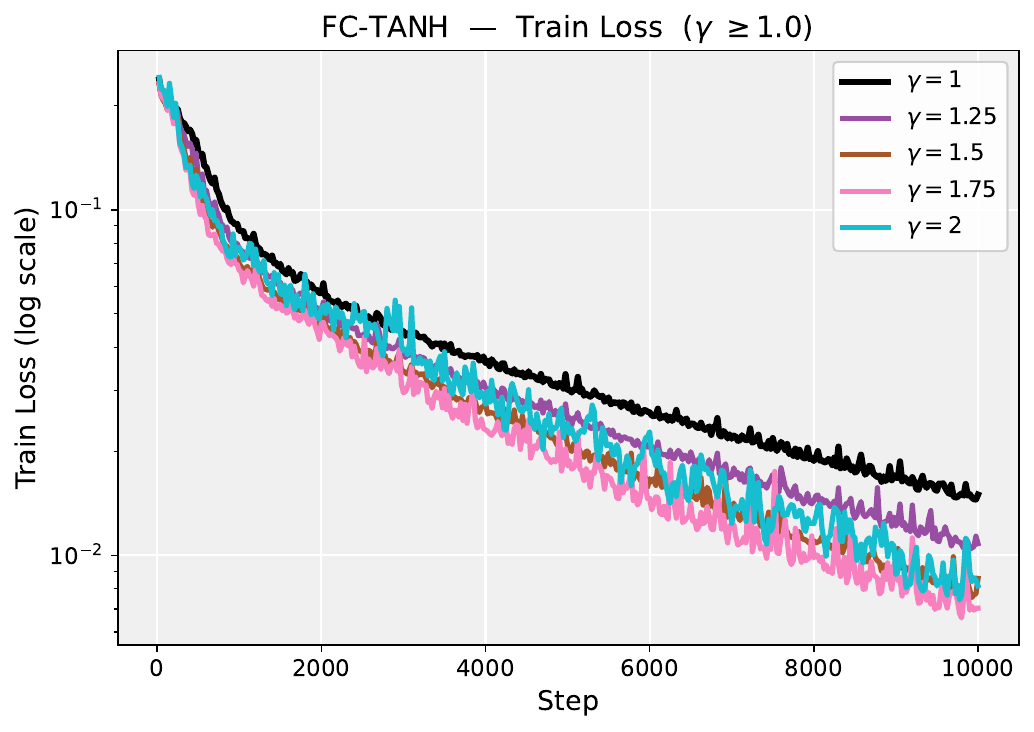}
    \end{minipage}
    \caption{FC TANH model trained on MNIST-5k. We fix $\alpha=1.0$ and vary $\gamma$.} 
\label{app:fig:fc_alpha_fix_gamma_sweep}
\end{figure}

\begin{figure}[H]
    \centering
    \begin{minipage}[b]{0.49\textwidth}
        \centering
        \includegraphics[width=\textwidth]{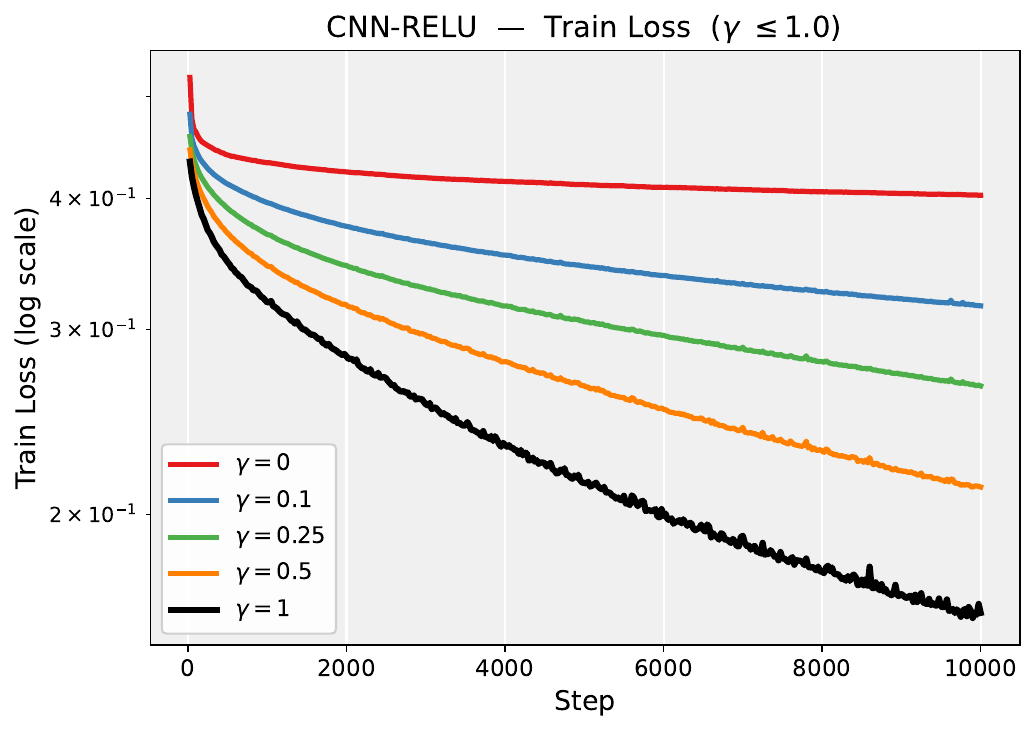}
    \end{minipage}
    \hfill
    \begin{minipage}[b]{0.49\textwidth}
        \centering
        \includegraphics[width=\textwidth]{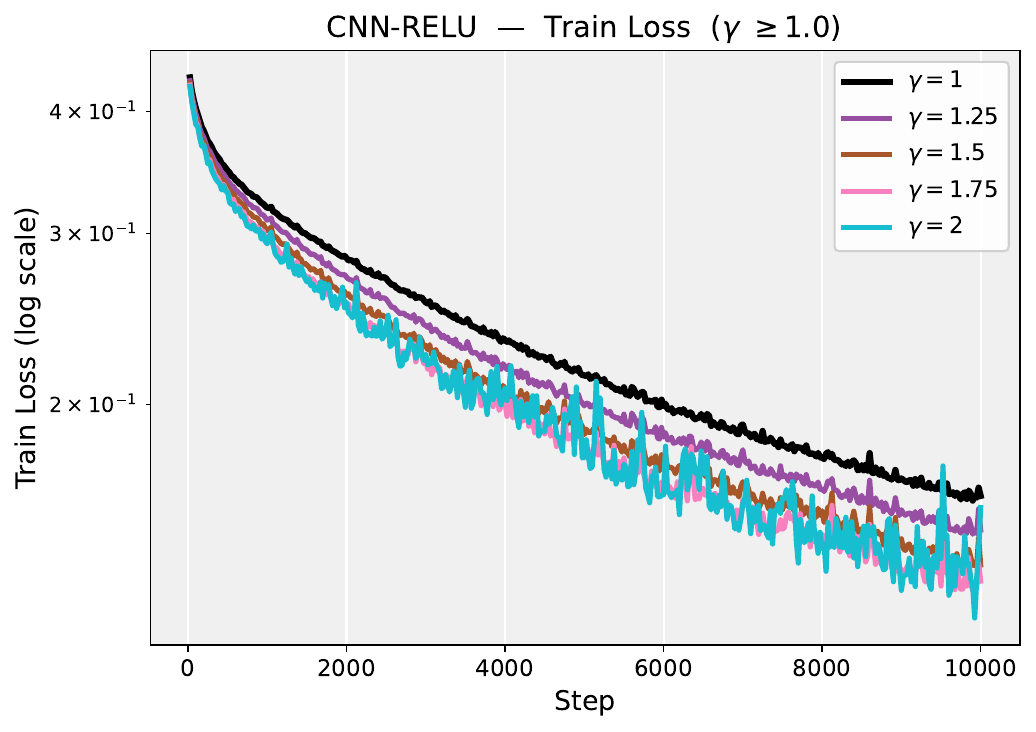}
    \end{minipage}
    \caption{CNN ReLU model trained on CIFAR10-5k. We fix $\alpha=1.0$ and vary $\gamma$.} 
\label{app:fig:cnn_alpha_fix_gamma_sweep}
\end{figure}

\begin{figure}[H]
    \centering
    \begin{minipage}[b]{0.49\textwidth}
        \centering
        \includegraphics[width=\textwidth]{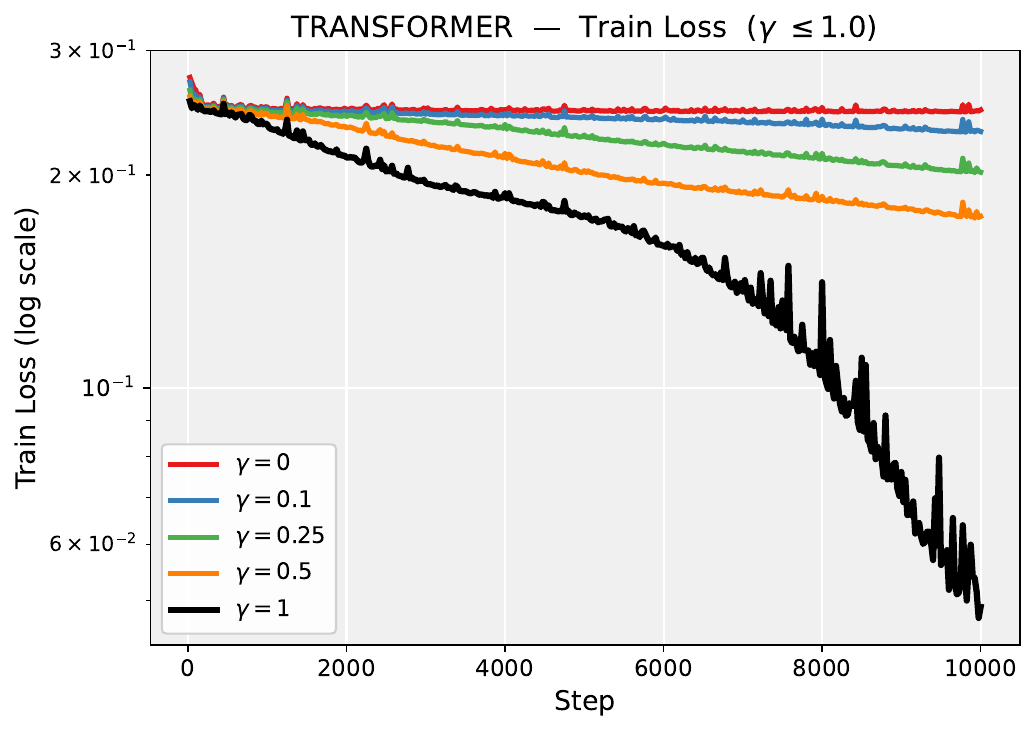}
    \end{minipage}
    \hfill
    \begin{minipage}[b]{0.49\textwidth}
        \centering
        \includegraphics[width=\textwidth]{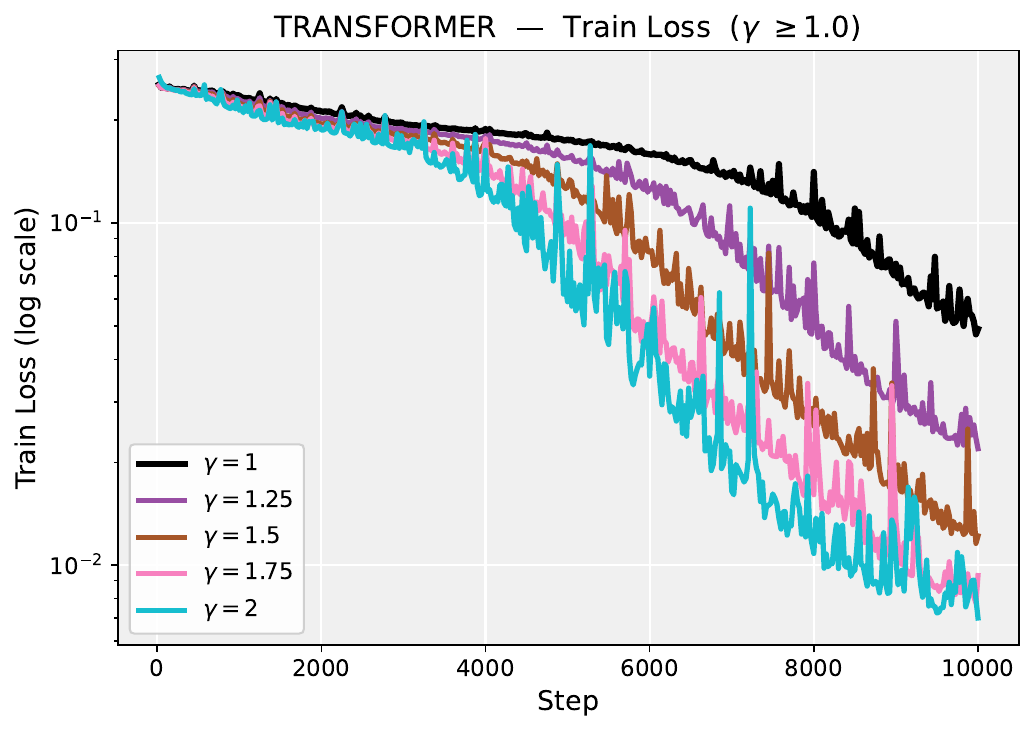}
    \end{minipage}
    \caption{Transformer model trained on SST2-5k. We fix $\alpha=1.0$ and vary $\gamma$.} 
\label{app:fig:transformer_alpha_fix_gamma_sweep}
\end{figure}

\paragraph{ii) Fixing \(\alpha=1.0\) and varying \(\gamma\).}
In this setting, we fix \(\alpha=1.0\) and sweep
\[
    \gamma \in
    \{0.0, 0.1, 0.25, 0.5, 1.0, 1.25, 1.5, 1.75, 2.0\}.
\]
When \(\gamma=1.0\), the filtered update reduces to standard Local SGD, and we
mark this baseline curve in black. We report the results in
Figures~\ref{app:fig:fc_alpha_fix_gamma_sweep},
\ref{app:fig:cnn_alpha_fix_gamma_sweep}, and
\ref{app:fig:transformer_alpha_fix_gamma_sweep}. In each figure, the left panel
shows the training loss curves for \(\gamma \leq 1.0\), corresponding to
suppression of the component orthogonal to the gap-estimated dominant subspace,
while the right panel shows the curves for \(\gamma \geq 1.0\), corresponding
to amplification of this orthogonal, bulk component.

As can be observed in Figures~\ref{app:fig:fc_alpha_fix_gamma_sweep},
\ref{app:fig:cnn_alpha_fix_gamma_sweep}, and
\ref{app:fig:transformer_alpha_fix_gamma_sweep}, suppressing the estimated bulk component
by setting \(\gamma<1.0\) substantially slows down optimization and leads to a
much higher final training loss, across all three settings. In particular, small values of \(\gamma\)
prevent the training loss from decreasing effectively, suggesting that the
orthogonal component \(p_{\mathrm{bulk}}^c\) carries much of the useful descent
signal. Conversely, amplifying this component with \(\gamma>1.0\) consistently
accelerates optimization and reaches a significantly lower final training loss
than standard Local SGD. The improvement is especially clear in the Transformer
experiment, where larger \(\gamma\) values lead to a much faster late-stage
drop in training loss. At the same time, very large values of \(\gamma\) can
introduce additional instability, as seen from the noisier loss curves for the
largest amplification factors. Overall, these results complement the
\(\alpha\)-sweep: suppressing the gap-estimated dominant component and
amplifying the orthogonal bulk component both improve the rate at which the
training loss decreases.

\newpage

\vspace{-3mm}

\vspace{-3mm}

\vspace{-3mm}

\vspace{-3mm}
